\pdfoutput=1
\documentclass[11pt]{article}

\usepackage[]{acl}

\usepackage{times}
\usepackage{latexsym}

\usepackage{times}
\usepackage{latexsym}
\usepackage{caption}
\usepackage{subcaption}
\usepackage[T1]{fontenc}
\usepackage[utf8]{inputenc}

\usepackage{microtype}
\usepackage{appendix}
\usepackage{amsmath}
\usepackage{amssymb}
\usepackage{array,multirow,graphicx}
\usepackage{floatrow}
\usepackage[T1]{fontenc}
\usepackage{color}
\usepackage{colortbl}
\usepackage{soul}
\colorlet{soulred}{orange!30}
\sethlcolor{soulred}
\usepackage{microtype}

\author{George Chrysostomou \quad \quad Nikolaos Aletras\\
  Department of Computer Science, University of Sheffield \\
  United Kingdom \\
  \texttt{\{gchrysostomou1, n.aletras\}@sheffield.ac.uk} \\}

\title{An Empirical Study on Explanations in Out-of-Domain Settings}

\newcommand{\bert}{\texttt{BERT}}

\begin{document}
\maketitle

\begin{abstract}
   %The increasing adoption of large pre-trained models in safety-critical domains, such as healthcare, has expanded the need for explaining their predictions. This led to a surge in developing approaches that generate explanations, either via identifying the most important tokens in the input (i.e. through feature attributions) or by designing inherently faithful models.
   Recent work in Natural Language Processing has focused on developing approaches that extract faithful explanations, either via identifying the most important tokens in the input (i.e. post-hoc explanations) or by designing inherently faithful models that first select the most important tokens and then use them to predict the correct label (i.e. select-then-predict models).
   Currently, these approaches are largely evaluated on in-domain settings. Yet, little is known about how post-hoc explanations and inherently faithful models perform in out-of-domain settings.
%   Recent work in natural language processing on  evaluating the faithfulness of model explanations (i.e. how accurately it represents a model's true reasoning behind a prediction) typically focuses on in-domain settings.
%   Yet, little is known about how faithful explanations are in out-of-domain settings. %This can have potential implications when explanations are used for sensitive tasks, such as automatic fact-checking on social media.
    In this paper, we conduct an extensive empirical study that examines: (1) the out-of-domain faithfulness of post-hoc explanations, generated by five feature attribution methods; and (2) the out-of-domain performance of two inherently faithful models over six datasets. 
    % Similar to model predictive performance, we expect degradation in post-hoc explanation faithfulness and select-then-predict performance when on out-of-domain settings.
    Contrary to our expectations, results show that in many cases out-of-domain post-hoc explanation faithfulness measured by sufficiency and comprehensiveness is higher compared to in-domain.
    We find this misleading and suggest using a random baseline as a yardstick for evaluating post-hoc explanation faithfulness. 
    Our findings also show that select-then predict models demonstrate comparable predictive performance in out-of-domain settings to full-text trained models.\footnote{Code available at: \url{https://github.com/GChrysostomou/ood_faith}}% whilst having the added benefit of offering inherently faithful rationales. 
    % We find that using metrics such as sufficiency and comprehensiveness may be misleading on measuring the actual faithfulness of out-of-domain post-hoc explanations and propose an alternative approach to interpreting results. Our findings also show that select-then predict models, demonstrate comparable predictive performance in out-of-domain settings to full-text trained models, whilst having the added benefit of offering inherently faithful rationales. 
    % We also show that using inherently faithful select-then predict models, results in high predictive performance in out-of-domain settings, often being comparable to their full-text trained model counterparts. 
\end{abstract}

\section{Introduction}

An explanation or rationale\footnote{We use these terms interchangeably throughout our work.}, typically consists of a subset of the input that contributes more to the prediction. Extracting faithful explanations is important for studying model behavior \citep{adebayo-nips} and assisting in tasks requiring human decision making, such as clinical text classification \citep{chakrabarty-etal-2019-pay}, misinformation detection \citep{popat-etal-2018-declare,mu2020} and legal text classification~\citep{chalkidis-etal-2019-neural,chalkidis-etal-2021-paragraph}. A faithful explanation is one which accurately represents the reasoning behind a model's prediction \citep{jacovi-goldberg-2020-towards}

\begin{figure*}[!t]
     \centering
     \begin{subfigure}[b]{0.45\textwidth}
         \centering
         \includegraphics[width=\textwidth]{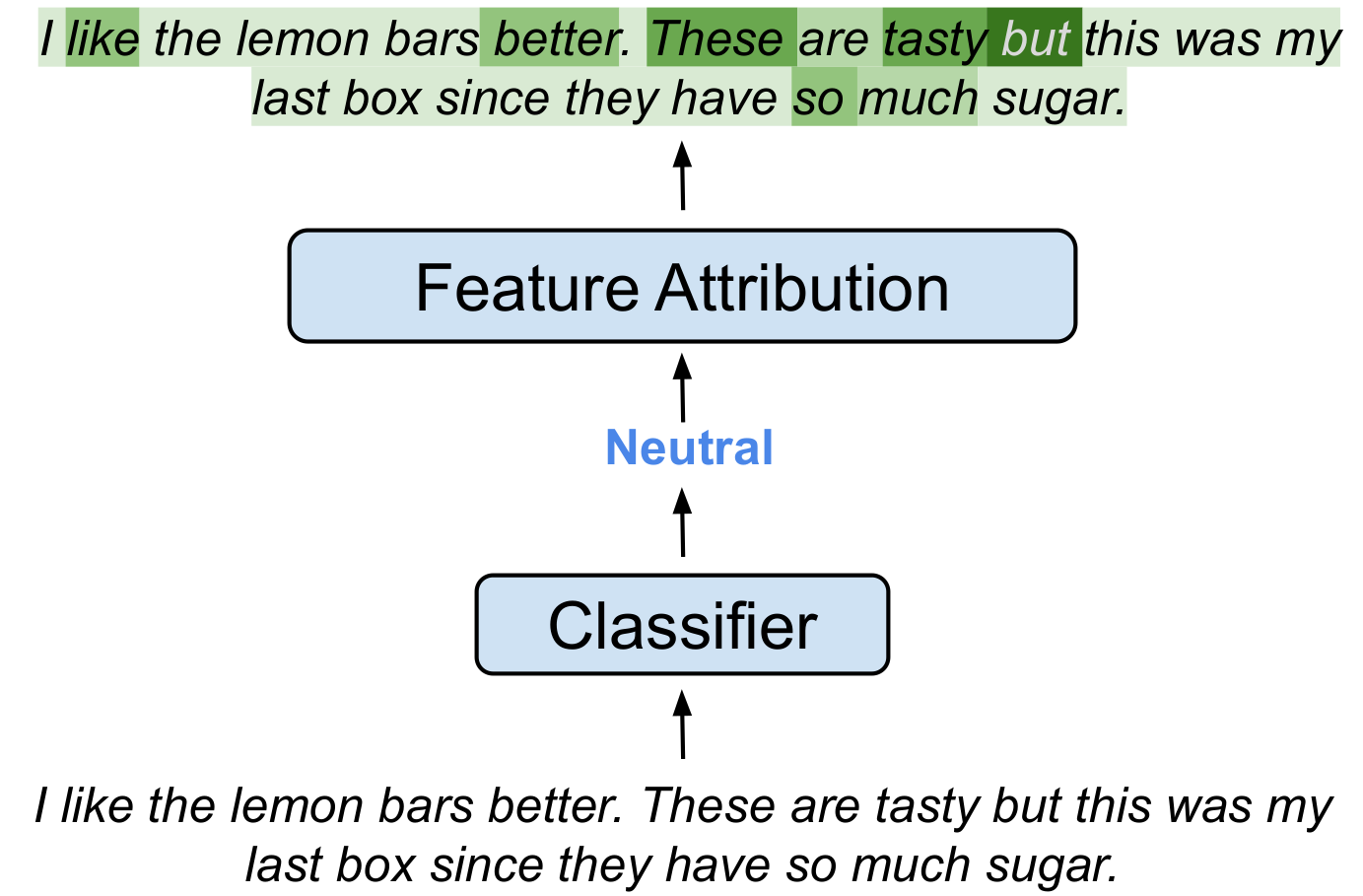}
         \caption{Post-hoc explanation}
     \end{subfigure}
     \hspace{0.0em}
     \begin{subfigure}[b]{0.45\textwidth}
         \centering
         \includegraphics[width=\textwidth]{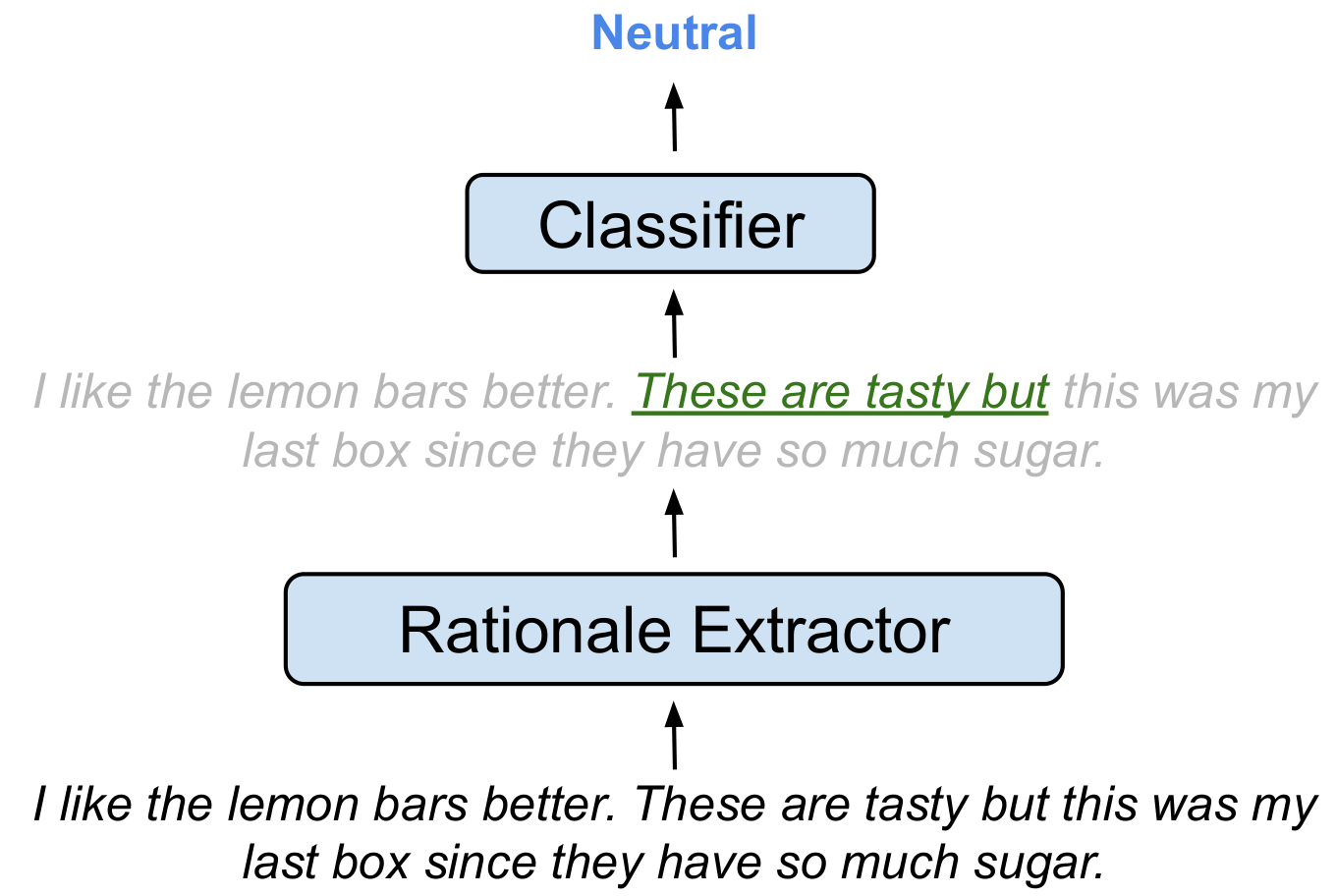}
         \caption{Select-then-predict model}
     \end{subfigure}
     \caption{An example of rationale extraction using: (a) a feature attribution approach to identify the most important subset of the input (post-hoc explanation); and (b) using inherently faithful, select-then-predict models.}
        
        \label{fig:examples}
    
\end{figure*}

Two popular methods for extracting explanations are through feature attribution approaches (i.e. \emph{post-hoc} explanation methods) or via inherently faithful classifiers (i.e. \emph{select-then-predict} models). The first computes the contribution of different parts of the input with respect to a model's prediction \citep{integrated_gradients, ribeiro2016model, shrikumar-deeplift}. The latter consists of using a rationale extractor to identify the most important parts of the input and a rationale classifier, a model trained using as input only the extractor's rationales \citep{bastings-etal-2019-interpretable,jain-etal-2020-learning, guerreiro2021spectra}.\footnote{We refer to the rationale generator (i.e. generating a rationale mask) from \citet{bastings-etal-2019-interpretable} and \citet{jain-etal-2020-learning} as a rationale extractor, to avoid any confusion between these approaches and free-text rationales \citep{wiegreffe-etal-2021-measuring}.} Figure \ref{fig:examples} illustrates the two approaches with an example.
% For select-then-predict models the generator and classifier can be trained jointly \citep{lei-etal-2016-rationalizing, bastings-etal-2019-interpretable} or independently \citep{treviso-martins-2020-explanation, jain-etal-2020-learning}.

Currently, these explanation methods have been mostly evaluated on in-domain settings (i.e. the train and test data come from the same distribution). However, when deploying models in real-world applications, inference might be performed on data from a different distribution, i.e. out-of-domain \citep{desai-durrett-2020-calibration, NEURIPS2019_8558cb40}. This can create implications when extracted explanations (either using post-hoc methods or through select-then-predict models) are used for assisting human decision making. Whilst we are aware of the limitations of current state-of-the-art models in out-of-domain predictive performance \citep{hendrycks-etal-2020-pretrained}, to the best of our knowledge, how faithful out-of-domain post-hoc explanations are has yet to be explored. Similarly, we are not aware how inherently faithful select-then-predict models generalize in out-of-domain settings.
% Currently, post-hoc explanation faithfulness and select-then-predict performance have been only evaluated in in-domain settings (i.e. train and test data come from the same distribution). However when deploying models in real-world applications, inference may be performed on data with different distribution (out-of-domain) \citep{desai-durrett-2020-calibration, NEURIPS2019_8558cb40}. This creates implications when model explanations are used for assisting human decision making. Whilst we are aware of the limitations of current state-of-the-art models in out-of-domain predictive performance \citep{hendrycks-etal-2020-pretrained}, to the best of our knowledge, how faithful explanations are in out-of-domain settings has yet to be explored. 

Inspired by this, we conduct an extensive empirical study to examine the faithfulness of five feature attribution approaches and the generalizability of two select-then-predict models in out-of-domain settings across six dataset pairs. We hypothesize that similar to model predictive performance, post-hoc explanation faithfulness reduces in out-of-domain settings and that select-then-predict performance degrades. %However, our findings do not agree in all cases with our hypothesis. 
Our contributions are as follows: %Our results counter-intuitively show that the faithfulness of explanations extracted using post-hoc methods \emph{improves} out-of-domain compared to in-domain for standard metrics such as sufficiency and comprehensiveness. We provide an analysis for this phenomenon and propose an alternative approach to interpreting results in \S\ref{sec:Comparing_the_faithfulness_in_feature_attribution}.
%On the contrary, the select-then-predict models exhibit lower predictive out-of-domain performance compared to in-domain as expected. Moreover, their out-of-domain performance is often comparable to those of full-text models (see \S\ref{sec:results_select_predict}), suggesting that they also achieve inherent faithfulness without substantial sacrifices in predictive performance. To summarize our contributions are:
\begin{itemize}
    \item To the best of our knowledge, we are the first to assess the faithfulness of post-hoc explanations and performance of select-then-predict models in out-of-domain settings.
    \item We show that post-hoc explanation sufficiency and comprehensiveness show misleading increases in out-of-domain settings. We argue that they should be evaluated alongside a random baseline as yardstick out-of-domain. 
    % \item We show that using sufficiency and comprehensiveness to quantify faithfulness out-of-domain can yield misleadingly high scores in certain cases. and propose an alternative approach to interpreting results.
    \item We demonstrate that select-then-predict classifiers can be used in out-of-domain settings. They lead to comparable predictive performance to models trained on full-text, whilst offering inherent faithfulness.
\end{itemize}

\section{Related Work}

\subsection{Rationale Extraction
\label{sec:generating_explanations}}

Given a model $\mathcal{M}$, we are interested in explaining why $\mathcal{M}$ predicted $\hat{y}$ for a particular instance $\mathbf{x} \in \mathbf{X}$. An extracted rationale $\mathcal{R}$, should therefore represent as accurately as possible the most important subset of the input ($\mathcal{R} \in \mathbf{x}$) which contributed mostly towards the model's prediction $\hat{y}$.

Currently, there are two popular approaches for extracting rationales. The first consists of using feature attribution methods that attribute to the input tokens an importance score (i.e. how important an input token is to a model's $\mathcal{M}$ prediction $\hat{y}$). We can then form a rationale $\mathcal{R}$, by selecting the $K$ most important tokens (independent or contiguous) as indicated by the feature attribution method. The second select-then-predict approach focuses on training inherently faithful classifiers by jointly training two modules, a \emph{rationale extractor} and a \emph{rationale classifier}, trained only on rationales produced by the extractor \citep{lei-etal-2016-rationalizing, bastings-etal-2019-interpretable, treviso-martins-2020-explanation, jain-etal-2020-learning, guerreiro2021spectra}. Recent studies have used feature attribution approaches as part of the rationale extractor \citep{jain-etal-2020-learning, treviso-martins-2020-explanation}, showing improved classifier predictive performance. 

\subsection{Evaluating Rationale Faithfulness}
\label{sec:rationale_evaluation}

Having extracted $\mathcal{R}$, we need to evaluate the quality of the explanation (i.e. how faithful that explanation is for a model's prediction). Typically, post-hoc explanations from feature attribution approaches are evaluated using input erasure \citep{serrano-smith-2019-attention, atanasova2020diagnostic, madsen2021evaluating}. This approach masks segments of the input to observe if the model's prediction changed. \citet{deyoung-etal-2020-eraser} proposed measuring the comprehensiveness and sufficiency of rationales as faithfulness metrics. A comprehensive rationale is one which is influential to a model's prediction, while a sufficient rationale that which is adequate for a model's prediction \citep{deyoung-etal-2020-eraser}. The term \textit{fidelity} is also used for jointly referring to comprehensiveness and sufficiency \citep{carton-etal-2020-evaluating}.
%has also used input erasure to evaluate the comprehensiveness and sufficiency of rationales as faithfulness metrics. 
\citet{carton-etal-2020-evaluating} suggested normalizing these metrics using the predictions of the model with a baseline input (i.e. an all zero embedding vector), to account for baseline model behavior.
Select-then-predict models are inherently faithful, as their classification component is trained only on extracted rationales \citep{jain-etal-2020-learning}. A good measure for measuring rationale quality is by evaluating the predictive performance of the classifier trained only on the rationales \cite{jain-etal-2020-learning, treviso-martins-2020-explanation}. A higher score entails that the extracted rationales are better when compared to those of a classifier with lower predictive performance.

\subsection{Explainability in Out-of-Domain Settings}

Given model $\mathcal{M}$ trained on an end-task, we typically evaluate its out-of-domain predictive performance on a test-set that does not belong to the same distribution as the data it was trained on \citep{hendrycks-etal-2020-pretrained}. Similarly, the model can also extract explanations $\mathcal{R}$ for its out-of-domain predictions. 

\citet{NEURIPS2018_4c7a167b} studied whether generating explanations for language inference match human annotations (i.e. plausible explanations). They showed that this is challenging in-domain and becomes more challenging in out-of-domain settings. In a similar direction, \citet{rajani-etal-2019-explain} and \citet{kumar-talukdar-2020-nile} examined model generated explanations in out-of-domain settings and find that explanation plausibility degrades compared to in-domain. 
\citet{kennedy-etal-2020-contextualizing} proposed a method for detecting model bias towards group identity terms using a post-hoc feature attribution approach. Then, they use them for regularizing models to improve out-of-domain predictive performance. 
\citet{adebayo-nips} have studied feature attribution approaches for identifying out-of-distribution images. They find that importance allocation in out-of-domain settings is similar to that of an in-domain model and thus cannot be used to detect such images. \citet{feder2021causal} finally argued that explanations can lead to errors in out-of-distribution settings, as they may latch onto spurious features from the training distribution. 

These studies indicate that there is an increasing need for evaluating post-hoc explanation faithfulness and select-then-predict performance in out-of-domain settings. To the best of our knowledge, we are the first to examine these. 

% % Transformer-based models have been shown to be robust in out-of-distribution settings, often producing highly calibrated probability scores and outperforming other model architectures \citet{hendrycks-etal-2020-pretrained}. 

% ** WILL DISCUSS ABOUT PERFORMANCE AND PERHAPS THE ABEDAYO PAPER FOR USING THEM FOR MODEL DEBUGGING IN IMAGES **

% \citet{hendrycks-etal-2020-pretrained} show that pre-trained transformers improve out-of-distribution robustness. \citet{adebayo-nips} check if explanations can be used for model debugging. \citet{feder2021causal} discuss about causal inference and about explanations and the difference between correlation and causality and also about the need of evaluating explanations in out-of-distribution settings.

% ** IMPORTANT **
% \citet{arora2021types} talk about types of OOD distribution and how to detect them.
% ****

% \citet{xu-etal-2021-unsupervised} do out of domain unsupervised detection (i.e. they detect inputs that do not belong in the feature space of what the model has learned)
% SPECTRA: Sparse Structured Text Rationalization

\section{Extracting Rationales}

\subsection{Post-hoc Explanations \label{para:feature_attribution}} We employ a pre-trained \bert{}-base and fine-tune it on in-domain training data. We then extract post-hoc rationales for both the in-domain test-set and two out-of-domain test-sets. We compute input importance using five feature scoring methods and a random baseline:

\begin{itemize}
\item \textbf{Random (\textsc{Rand}): } Random allocation of token importance. 
\item\textbf{Attention ($\boldsymbol{\alpha}$):} Token importance corresponding to normalized attention scores \cite{jain-etal-2020-learning}.
\item\textbf{Scaled Attention ($\boldsymbol{\alpha} \nabla \boldsymbol{\alpha}$):} Attention scores $\alpha_i$ scaled by their corresponding gradients $\nabla \alpha_i= \frac{\partial \hat{y}}{\partial \alpha_i}$ \cite{serrano-smith-2019-attention}.
\item\textbf{InputXGrad ($\mathbf{x} \nabla \mathbf{x}$):} Attributes input importance by multiplying the input with its gradient computed with respect to the predicted class, where $\nabla x_i = \frac{\partial \hat{y}}{\partial x_i}$ \cite{kindermans2016investigating, atanasova2020diagnostic}.

\item\textbf{\textbf{Integrated Gradients ($\mathbf{IG}$)}:} Ranking words by computing the integral of the gradients taken along a straight path from a baseline input (zero embedding vector) to the original input \cite{integrated_gradients}. 

\item\textbf{\textbf{DeepLift}:} Ranking words according to the difference between the activation of each neuron and a reference activation (zero embedding vector) \cite{shrikumar-deeplift}. 

% \item\textbf{\textbf{LIME}:} Ranking words by learning an interpretable
% model locally around the prediction \cite{ribeiro2016model}.
\end{itemize}

\subsection{Select-then-Predict Models \label{par:select_then_predict}} 
We use two select-then-predict models: 

\begin{itemize}
    \item \textbf{HardKuma: } An end-to-end trained model, where the rationale extractor uses Hard Kumaraswamy variables to produce a rationale mask \textbf{z}, which the classifier uses to mask the input \citep{bastings-etal-2019-interpretable}. Model training takes advantage of reparameterized gradients compared to \texttt{REINFORCE} style training employed by \citet{lei-etal-2016-rationalizing} and has shown improved performance \citep{guerreiro2021spectra}. 
     \item \textbf{FRESH: }  We compute the predictive performance of a classifier trained on rationales extracted with feature attribution metrics (see $\S$\ref{para:feature_attribution}) using \textsc{FRESH}, following a similar approach to \citet{jain-etal-2020-learning}. We extract rationales from an extractor by (1) selecting the top-$k$ most important tokens (\textsc{TopK}) and (2) selecting the span of length $k$ with the highest overall importance (\textsc{Contiguous}).
\end{itemize}

We use \bert{}-base for the extraction and classification components of FRESH similar to \citet{jain-etal-2020-learning}. However, for HardKuma we opt using a bi-\texttt{LSTM} \citep{hochreiter1997long} as it provides comparable or improved performance over \bert{} variants \citep{guerreiro2021spectra}, even after hyperparameter tuning.\footnote{See model details and hyper-parameters in Appendix \ref{app:extended_models}}

\section{Experimental Setup}

\subsection{Datasets}

For evaluating out-of-domain model explanation, we consider the following datasets (see Table \ref{tab:data_characteristic} and Appendix \ref{app:extended_data} for details):

\paragraph{SST:} Stanford Sentiment Treebank (SST) consists of sentences tagged with sentiment on a 5-point-scale from negative to positive \citep{socher-etal-2013-recursive}. We remove sentences with neutral sentiment and label the remaining sentences as negative or positive if they have a score lower or higher than 3 respectively \citep{jain-wallace-2019-attention}.

\paragraph{IMDB:} The Large Movie Reviews Corpus consists of movie reviews labeled either as positive or negative \citep{maas-etal-2011-learning, jain-wallace-2019-attention}. 

\paragraph{Yelp:} Yelp polarity review texts. Similar to \citet{NIPS2015_250cf8b5} we construct a binary classification task to predict a polarity label by considering one and two stars as negative, and three and four stars as positive.

\paragraph{Amazon Reviews:} We form 3-way classification tasks by predicting the sentiment (negative, neutral, positive) of Amazon product reviews across 3 item categories: (1) Digital Music (\textbf{AmazDigiMu}); (2) Pantry (\textbf{AmazPantry}); and (3) Musical Instruments (\textbf{AmazInstr}) \citep{ni-etal-2019-justifying}.

\renewcommand{\arraystretch}{1.2}
\begin{table}[!t]
    \centering
    \small
    \begin{tabular}{l||cc}
        \textbf{Dataset} & \textbf{C} & \textbf{Splits}\\ \hline \hline
        SST & 2 & 6,920 / 872 / 1,821  \\
        IMDB & 2 & 20,000 / 2,500 / 2,500 \\
        Yelp & 2 & 476,000 / 84,000 / 38,000 \\
        AmazDigiMu & 3 & 122,552 / 21,627 / 25,444  \\
        AmazPantry & 3 & 99,423/ 17,546 / 20,642 \\
        AmazInstr & 3 & 167,145 / 29,497 / 34,702  \\
    \end{tabular}
    \caption{Dataset statistics with number of classes (\textbf{C}) and train/development/test \textbf{splits}. For more details see Appendix \ref{app:extended_data}.}
    \label{tab:data_characteristic}
\end{table}

\subsection{Evaluating Out-of-Domain Explanations}

% To evaluate the quality (i.e. faithfulness) of rationales in OOD settings we employ a similar setup to \citet{hendrycks-etal-2020-pretrained}. We first train a model $\mathcal{M}$ (either using the full text; an inherently faithful model; or a classifier trained on rationales) on ID data $\mathcal{D}_{ID}$. We then evaluate the model's performance in zero-shot fashion (i.e. we do not retrain the model) on two OOD datasets $\mathcal{D}_{OOD-1}$ and $\mathcal{D}_{OOD-2}$. We compare ID and OOD faithfulness by:

% \paragraph{Feature Attribution:} We use a  fine-tuned  \bert{} (as model $\mathcal{M}$) on in-domain training data and extract importance scores using feature attribution approaches for the in- and out-of-domain test-set instances. We measure explanation faithfulness for feature attribution using:
\paragraph{Post-hoc Explanations:}

%Similar to model predictive performance, we expect a deterioration in post-hoc explanation faithfulness on the out-of-domain test-sets when comparing to the in-domain test-set. 
We evaluate post-hoc explanations using:

\renewcommand{\arraystretch}{1.1}
\begin{table*}[!t]
    \centering
    \small
    \setlength{\tabcolsep}{2.2pt}
    \begin{tabular}{cc||c||cccccc||cccccc}
         \textbf{Train} & \textbf{Test} & \textbf{Full-text} &  \multicolumn{6}{c||}{\textbf{Normalized Sufficiency}} & \multicolumn{6}{c}{\textbf{Normalized Comprehensiveness}}\\
          &  & \textbf{F1} & Rand & $\alpha\nabla\alpha$ & $\alpha$ &  DeepLift &       $x\nabla x $ &        IG  & Rand & $\alpha\nabla\alpha$ & $\alpha$ &  DeepLift &       $x\nabla x $ &        IG  \\\hline \hline
        \multirow{3}{*}{SST} &   SST & 90.1  & 0.38 &             0.51 &      0.42 &     0.42 &       0.40 &  0.41&  0.19 &             0.39 &      0.22 &     0.25 &      0.26 &  0.26  \\ 
              &        IMDB & 84.3  & 0.31 &             0.53 &      0.39 &     0.32 &      0.31 &  0.32 &  0.23 &             0.54 &      0.34 &     0.27 &      0.27 &  0.28  \\
              &        Yelp & 87.9  & 0.32 &             0.56 &       0.40 &     0.35 &      0.33 &  0.34 &  0.21 &             0.48 &      0.28 &     0.24 &      0.24 &  0.25  \\\hline 
        \multirow{3}{*}{IMDB}  & IMDB & 91.1  & 0.32 &             0.55 &      0.46 &     0.36 &      0.36 &  0.36 & 0.16 &             0.48 &      0.31 &     0.25 &      0.23 &  0.24  \\
             &         SST & 85.8  & 0.24 &             0.35 &      0.28 &     0.28 &      0.27 &  0.27 &  0.27 &             0.46 &      0.32 &     0.33 &      0.33 &  0.33  \\
             &        Yelp &  91.0  & 0.35 &             0.48 &      0.41 &     0.36 &      0.36 &  0.36 &  0.21 &             0.45 &      0.32 &     0.26 &      0.26 &  0.26  \\   \hline
         \multirow{3}{*}{Yelp} &  Yelp & 96.9  & 0.23 &             0.32 &      0.31 &     0.29 &      0.24 &  0.25  &  0.12 &              0.20 &      0.14 &     0.16 &      0.15 &  0.16 \\
             &         SST &  86.8  & 0.41 &             0.45 &      0.43 &     0.44 &      0.41 &  0.41 & 0.17 &             0.24 &      0.18 &     0.21 &      0.22 &  0.22  \\
             &        IMDB &   88.6  &  0.18 &             0.34 &      0.32 &     0.25 &      0.22 &  0.22  &  0.19 &             0.34 &      0.29 &     0.23 &      0.23 &  0.24 \\ \hline \hline
        \multirow{3}{*}{AmazDigiMu} &  AmazDigiMu &   70.6  &  0.34 &             0.56 &      0.34 &     0.31 &      0.41 &  0.39 & 0.13 &             0.32 &      0.14 &      0.10 &      0.16 &  0.17  \\
       &   AmazInstr &  61.2  & 0.29 &             0.54 &      0.32 &     0.31 &      0.33 &  0.32 &   0.19 &             0.47 &      0.23 &     0.19 &      0.22 &  0.23  \\
       &  AmazPantry & 64.6  & 0.33 &             0.55 &      0.33 &     0.31 &      0.37 &  0.36  &   0.21 &             0.46 &      0.22 &     0.17 &      0.23 &  0.25 \\   \hline
        \multirow{3}{*}{AmazPantry} & AmazPantry &  70.2  & 0.25 &             0.46 &      0.36 &     0.19 &      0.28 &  0.27  &    0.20 &             0.42 &      0.31 &     0.15 &      0.25 &  0.25  \\
       &  AmazDigiMu & 59.5  & 0.24 &             0.47 &      0.37 &     0.19 &      0.27 &  0.26 &   0.19 &             0.41 &      0.32 &     0.15 &      0.23 &  0.24  \\
       &   AmazInstr &   64.5  & 0.17 &             0.42 &       0.30 &     0.15 &       0.20 &   0.20 &  0.24 &             0.52 &       0.40 &     0.23 &       0.30 &   0.30  \\ \hline
        \multirow{3}{*}{AmazInstr} &  AmazInstr & 71.5  &  0.16 &             0.34 &      0.18 &     0.21 &      0.18 &  0.17  &  0.26 &             0.52 &      0.26 &     0.29 &      0.28 &  0.29 \\
        &  AmazDigiMu & 61.3  & 0.21 &             0.38 &      0.21 &     0.22 &      0.24 &  0.22 &   0.23 &             0.46 &       0.20 &     0.22 &      0.24 &  0.25  \\
        &  AmazPantry & 68.2  &  0.22 &             0.39 &      0.21 &     0.23 &      0.24 &  0.23 &   0.27 &             0.51 &      0.22 &     0.25 &      0.27 &  0.29  \\
        
    \end{tabular}
    \caption{AOPC Normalized Sufficiency and Comprehensiveness (higher is better) in-domain and out-of-domain for five feature attribution approaches and a random attribution baseline. %For reference, we also include the F1-macro predictive performance of models trained on the full-text averaged across 3 runs with standard deviation.
    }
    \label{tab:faithfulness_feature_scoring_comp-aopc}
\end{table*}

\begin{itemize}
    \item  \textbf{Normalized Sufficiency (NormSuff)} measures the degree to which the extracted rationales are adequate for a model to make a prediction \cite{deyoung-etal-2020-eraser}. Following \citet{carton-etal-2020-evaluating}, we bind sufficiency between 0 and 1 and use the reverse difference so that higher is better: 
    \begin{equation}
    \small
    \begin{aligned}
        \text{Suff}(\mathbf{x}, \hat{y}, \mathcal{R}) = 1 - max(0, p(\hat{y}| \mathbf{x})- p (\hat{y}|\mathcal{R}))\\
        \text{NormSuff}(\mathbf{x}, \hat{y}, \mathcal{R}) = \frac{\text{Suff}(\mathbf{x}, \hat{y}, \mathcal{R}) - \text{Suff}(\mathbf{x}, \hat{y}, 0)}{1 - \text{Suff}(\mathbf{x}, \hat{y}, 0)}
     \end{aligned}
    \end{equation}
    \noindent where $\text{Suff}(\mathbf{x}, \hat{y}, 0)$ is the sufficiency of a baseline input (zeroed out sequence) and $\hat{y}$ the model predicted class using the full text $\mathbf{x}$ as input. 
    
    \item \textbf{Normalized Comprehensiveness (NormComp)} measures the influence of a rationale to a prediction \cite{deyoung-etal-2020-eraser}.  For an explanation to be highly comprehensive, the model's prediction after masking the rationale should have a large difference compared to the model's prediction using the full text. Similarly to \citet{carton-etal-2020-evaluating}, we bind this metric between 0 and 1 and normalize it:
    \begin{equation}
    \small
    \begin{aligned}
        \text{Comp}(\mathbf{x}, \hat{y}, \mathcal{R}) = max(0, p(\hat{y}| \mathbf{x})- p (\hat{y}|\mathbf{x}_{\backslash\mathcal{R}}))\\
        \text{NormComp}(\mathbf{x}, \hat{y}, \mathcal{R}) = \frac{\text{Comp}(\mathbf{x}, \hat{y}, \mathcal{R})}{1 - \text{Suff}(\mathbf{x}, \hat{y}, 0)}
    \end{aligned}
    \end{equation}
\end{itemize}

To measure sufficiency and comprehensiveness across different explanation lengths we compute the “Area Over the Perturbation Curve" (AOPC) following \citet{deyoung-etal-2020-eraser}. We therefore compute and report the average normalized sufficiency and comprehensiveness scores when keeping (for sufficiency) or masking (for comprehensiveness) the top 2\%, 10\%, 20\% and 50\% of tokens extracted by an importance attribution function.\footnote{We also present results for each of these rationale lengths in Appendix \ref{app:post-hoc-extended}. }

We omit from our evaluation the Remove-and-Retrain method \citep{madsen2021evaluating} as it requires model retraining. Whilst this could be applicable for in-domain experiments where retraining is important, in this work we evaluate explanation faithfulness in zero-shot out-of-domain settings.

\paragraph{Select-then-Predict Models:} We first train select-then-predict models in-domain and then measure their predictive performance on the in-domain test-set and on two out-of-domain test-sets \citep{jain-etal-2020-learning, guerreiro2021spectra}. Our out-of-domain evaluation is performed without re-training (zero-shot). Similar to full-text trained models, we expect that predictive performance deteriorates out-of-domain. However, we assume that explanations from a select-then-predict model should generalize better in out-of-domain settings when the predictive performance approaches that of the full-text trained model. 

We do not conduct human experiments to evaluate explanation faithfulness, since that is only relevant to explanation plausibility (i.e. how intuitive to humans a rationale is \cite{jacovi-goldberg-2020-towards}) and in practice faithfulness and plausibility do not correlate \cite{atanasova2020diagnostic}.

\section{Results}

\subsection{Post-hoc Explanation Faithfulness}
\label{sec:Comparing_the_faithfulness_in_feature_attribution}

Table \ref{tab:faithfulness_feature_scoring_comp-aopc} presents the normalized comprehensiveness and sufficiency scores for post-hoc explanations on in-domain and out-of-domain test-sets, using five feature attribution methods and a random baseline. For reference, we include the averaged F1 performance across 5 random seeds, of a \bert{}-base model finetuned on the full text and evaluated in- and out-of-domain (Full-text F1).\footnote{We report predictive performance for all models and standard deviations in the Appendix.}

In-domain results show that feature attribution performance varies largely across datasets. This is in line with the findings of \citet{atanasova2020diagnostic} and \citet{madsen2021evaluating} when masking rationales (i.e. comprehensiveness). We find the only exception to be $\alpha \nabla \alpha$, which consistently achieves the highest comprehensiveness and sufficiency scores across all in-domain datasets. For example $\alpha\nabla\alpha$ evaluated on in-domain AmazDigiMu, results in sufficiency of 0.56 compared to the second best of 0.39 with IG. 

Contrary to our expectations, results show that post-hoc explanation sufficiency and comprehensiveness are in many cases higher in out-of-domain test-sets compared to in-domain. For example using DeepLift, comprehensiveness for the in-domain test-set in Yelp (0.16) is lower compared to the out-of-domain test-sets (0.21 for SST and 0.23 for IMDB). This is also observed when measuring sufficiency with $\alpha\nabla\alpha$, scoring 0.32 when tested in-domain on Yelp and 0.45 for the out-of-domain SST test-set. 
% These results suggest that sufficiency and comprehensiveness can be misleading and should not be used standalone for evaluating post-hoc explanation faithfulness out-of-domain. We argue that they should be used in conjunction with a random baseline for measuring their relative difference. 
%As we observe increases for sufficiency and comprehensiveness of post-hoc explanations in out-of-domain settings, so do 

Apart from increased sufficiency and comprehensiveness scores in out-of-domain post-hoc explanations, we also observe increased scores obtained by our random baseline. 
In fact, the random baseline outperforms several feature attribution approaches in certain cases in out-of-domain settings . As an example, consider the case where the model has been trained on AmazInstr and tested on AmazPantry. Our random baseline achieves a comprehensiveness score of 0.27 while $\alpha$, DeepLift, $x\nabla x$ perform similarly or lower (0.22, 0.25 and 0.27 respectively). Similarly, using a model trained on Yelp and tested on SST, the random baseline produces equally sufficient rationales to $x \nabla x$ and IG, with all of them achieving 0.41 normalized sufficiency. 
A glaring exception to this pattern is $\alpha \nabla \alpha$, which consistently outperforms both the random baseline and all other feature attribution approaches in in- and out-of-domain settings, suggesting that it produces the more faithful explanations. For example with out-of-domain AmazPantry test data, using a model trained on AmazInstr results in sufficiency scores of 0.39 with $\alpha\nabla\alpha$. This is a 0.15 point increase compared to the second best ($x\nabla x$ with 0.24).

%Similar to \citet{madsen2021evaluating}, 
We recommend considering \emph{a feature attribution for producing faithful explanations out-of-domain, if it only scores above a baseline random attribution. We suggest that the higher the deviation from the random baseline, the more faithful an explanation is}. 

\renewcommand{\arraystretch}{1.1}
\begin{table}[!t]
    \setlength\tabcolsep{1pt}
    \centering
    % \scriptsize
    \small
    \begin{tabular}{l||cc|cc}

        \textbf{Train} & \textbf{Test}  & \textbf{Full-text} & \textbf{HardKuma} & \textbf{L} \\
         &   & \textbf{F1} & \textbf{F1} & \textbf{(\%)} \\ \hline \hline
         \multirow{3}{*}{SST} &         SST &    81.7  &           77.6  &  56.8  \\
                             &        IMDB &    71.9  &  \textbf{65.7}  &  39.5  \\
                             &        Yelp &    68.7  &  \textbf{67.7}  &  32.7 \\  \hline
       \multirow{3}{*}{IMDB} &        IMDB &    87.4  &           82.0  &    1.9  \\
                             &         SST &    77.5  &           73.6  &   16.8  \\
                             &        Yelp &    41.0  &   \textbf{47.2}  &    3.1  \\  \hline
       \multirow{3}{*}{Yelp} &        Yelp &    96.0  &           92.4  &    7.4  \\
                             &         SST &    80.4  &           72.4  &   14.1  \\
                             &        IMDB &    84.5  &           73.3  &    4.7  \\  \hline
 \multirow{3}{*}{AmazDigiMu} &  AmazDigiMu &    67.6  &           66.8  &   18.4  \\
                             &   AmazInstr &    54.2  &   \textbf{53.3}  &   25.8  \\
                             &  AmazPantry &    55.3  &   \textbf{54.7}  &   27.8  \\  \hline
 \multirow{3}{*}{AmazPantry} &  AmazPantry &    67.9  &           66.6  &   18.9  \\
                             &  AmazDigiMu &    50.9  &   \textbf{51.0}  &   11.2  \\
                             &   AmazInstr &    55.9  &   \textbf{57.4}  &   18.2  \\  \hline
  \multirow{3}{*}{AmazInstr} &   AmazInstr &    67.2  &   \textbf{66.7}  &   19.2  \\
                             &  AmazDigiMu &    54.3  &   \textbf{53.7}  &   13.9  \\
                             &  AmazPantry &    61.1  &   \textbf{59.5}  &   24.4  \\
    \end{tabular}
    \caption{F1 macro performance (five runs) for HardKuma models and the selected rationale length (L). \textbf{Bold} denotes no significant difference between HardKuma and Full-text (t-test; p $>$ 0.05). For clarity, we include F1 scores with standard deviations in Appendix \ref{app:exteded_hardkuma}.}
    \label{tab:F1_ece_hardkuma}
\end{table}

\renewcommand{\arraystretch}{1.1}
\begin{table*}[!t]
    \centering
    \small
    \begin{tabular}{l||ll|ccccc}
\textbf{ Train} &   \textbf{Test}& \textbf{Full-text} &  \textbf{$\boldsymbol \alpha \nabla \alpha$} &   \textbf{$\boldsymbol \alpha$} &    \textbf{DeepLift} &   \textbf{$\boldsymbol x \nabla x$} &          \textbf{IG} \\ \hline \hline
         \multirow{3}{*}{SST (20\%)} &         SST &   90.1  &          87.7  &          81.1  &          84.4  &  76.3  &         76.8  \\
                             &        IMDB &   84.3  &          81.8  &          52.6  &          64.0  &  55.0  &         56.3  \\
                             &        Yelp &   87.9  &  \textbf{88.1 } &          72.6  &          75.4  &  59.6  &  63.9  \\\hline 
       \multirow{3}{*}{IMDB (2\%)} &        IMDB &   91.1  &          87.9  &          80.4  &          87.2  &  59.8  &         59.7  \\
                             &         SST &   85.8  &          80.9  &          71.8  &          70.1  &  69.6  &         70.7  \\
                             &        Yelp &   91.0  &          87.8  &          82.0  &          79.4  &  69.0  &  69.1  \\\hline 
       \multirow{3}{*}{Yelp (10\%)} &        Yelp &   96.9  &          94.0  &          90.4  &          93.6  &  70.5  &         71.9  \\
                             &         SST &   86.8  &          59.3  &          69.8  &          67.2  &  67.7  &         69.3  \\
                             &        IMDB &   88.6  &          78.0  &          64.5  &          66.6  &  53.0  &  55.8   \\\hline
 \multirow{3}{*}{AmazDigiMu (20\%)} &  AmazDigiMu &   70.6  &          66.1  &          63.4  &  \textbf{65.8 } &  51.9  &         65.8  \\
                             &   AmazInstr &   61.2  &  \textbf{58.0 } &  \textbf{57.2 } &  \textbf{57.4 } &  46.0  &         57.2  \\
                             &  AmazPantry &   64.6  &          59.1  &          56.5  &          56.5  &  44.8  &  44.8   \\\hline
 \multirow{3}{*}{AmazPantry (20\%)} &  AmazPantry &   70.2  &          67.3  &          62.6  &          67.2  &  48.6  &         48.7  \\
                             &  AmazDigiMu &   59.5  &  \textbf{57.7 } &          54.6  &          56.2  &  41.2  &         \textbf{57.7 } \\
                             &   AmazInstr &   64.5  &  \textbf{63.8 } &          58.0  &  \textbf{63.6 } &  40.1  &  40.3  \\\hline 
  \multirow{3}{*}{AmazInstr (20\%)} &   AmazInstr &   71.5  &          69.8  &          62.1  &          69.7  &  45.6  &         48.6  \\
                             &  AmazDigiMu &   61.3  &  \textbf{60.0 } &          53.2  &          57.8  &  43.8  &         \textbf{60.0 } \\
                             &  AmazPantry &   68.2  &          64.5  &          56.3  &          63.1  &  44.6  &  47.6   \\
\end{tabular}
   \caption{Average F1 macro performance of FRESH models (five runs) with the a priori defined rationale length in the brackets. \textbf{Bold} denotes no significant difference between FRESH and Full-text (t-test; p $>$ 0.05). For clarity, we present F1 scores with standard deviations in Appendix \ref{app:extended_fresh}.}
    \label{tab:F1_ece_select-then-predict-FRESH}
\end{table*}

\subsection{Select-then-predict Model Performance
\label{sec:results_select_predict}}

\paragraph{HardKuma:} Table \ref{tab:F1_ece_hardkuma} presents the F1-macro performance of HardKuma models \citep{bastings-etal-2019-interpretable} and the average rationale lengths (the ratio of the selected tokens compared to the length of the entire sequence) selected by the model. For reference, we also include the predictive performance of a full-text trained bi-\texttt{LSTM}. Results are averaged across 5 runs including standard deviations in brackets.

As expected, predictive performance of HardKuma models degrades when evaluated on out-of-domain data.
Surprisingly, though, we find that their performance is not significantly different (t-test; p-value $>$ 0.05) to that of the full-text \texttt{LSTM} in 9 out of the 12 out-of-domain dataset pairs. 
For example, by evaluating the out-of-domain performance of a HardKuma model trained on AmazDigiMu on the AmazPantry test-set, we record on average a score of 54.3 F1 compared to 55.3 with an \texttt{LSTM} classifier trained on full text.
We also observe that HardKuma models trained on SST and IMDB generalize comparably to models trained on full-text when evaluated on Yelp, however the opposite does not apply. Our assumption is that HardKuma models trained on Yelp, learn more domain-specific information due to the large training corpus (when compared to training on IMDB and SST) so they fail to generalize well out-of-domain. 
% using a model trained on AmazDigiMu on AmazPantry results to an F1-macro of 54.3 with HardKuma compared to 55.3 with an \texttt{LSTM} classifier. 

Results also show, that \emph{the length of rationales selected by HardKuma models depend on the source domain, i.e. training HardKuma on a dataset which favors shorter rationales, leads to also selecting shorter rationales out-of-domain}. For example, in-domain test-set explanation lengths are on average 56.8\% of the full-text input length for SST. In comparison, training a model on Yelp and evaluating on SST results in rationale lengths of 14.1\%. We observe that \emph{in certain cases, HardKuma models maintain the number of words, not the ratio to the sequence} in out-of-domain settings. For example, in-domain Yelp test-set rationales are about 11 tokens long that is the similar to the length selected when evaluating on IMDB using a model trained on Yelp. This is also observed where in-domain AmazInstr test-set rationales are on average 5 tokens long, which is the same rationale length when evaluating on AmazDigiMu using a model trained on AmazInstr. 
% We therefore assume that \emph{training HardKuma on a dataset which favors shorter rationales, leads to also  shorter rationales in out-of-domain settings}

In general, our findings show that in the majority of cases, using HardKuma in out-of-domain data results to comparable performance with their full-text model counterparts. This suggests that \textit{HardKuma models can be used in out-of-domain settings, without significant sacrifices in predictive performance whilst also offering faithful rationales}.

\paragraph{FRESH:} Table \ref{tab:F1_ece_select-then-predict-FRESH} shows the averaged F1-macro performance across 5 random seeds for FRESH classifiers on in- and out-of-domain using TopK rationales.\footnote{For clarity we include standard deviations and Contiguous results in Appendix \ref{app:extended_fresh}}. We also include the a priori defined rationale length in parentheses and the predictive performance of the Full-Text model for reference.\footnote{When evaluating out-of-domain, we use the average rationale length of the dataset we evaluate on. This makes FRESH experiments comparable with those of HardKuma.} %For clarity, we include standard deviations 

We first observe that in-domain predictive performance varies across feature attribution approaches with attention-based metrics ($\alpha\nabla\alpha$, $\alpha$) outperforming the gradient-based ones ($x\nabla x$, IG), largely agreeing with \citet{jain-etal-2020-learning}. We also find that $\alpha \nabla \alpha$ and DeepLift are the feature attribution approaches that lead to the highest predictive performance across all datasets.

As we initially hypothesized, performance of FRESH generally degrades when testing on out-of-domain data similarly to the behavior of models trained using the full text. The only exceptions are when using $x \nabla x$ and IG in IMDB. We argue that this is due to these feature attribution methods not being able to identify the appropriate tokens relevant to the task using a rationale length 2\% of the original input. Increasing the rationale length to 20\% (SST) and 10\% (Yelp) also increases the performance. 
Results also suggest that $\alpha\nabla\alpha$ and DeepLift outperform the rest of the feature attributions, with $\alpha \nabla \alpha$ being the best performing one in the majority of cases. 
In fact when using $\alpha \nabla \alpha$ or DeepLift, the out-of-domain performance of FRESH is not significantly different to that of models trained on full text (t-test; p-value $>$ 0.05) in 5 cases. For example, a FRESH model trained on AmazPantry and evaluated on AmazInstr records 63.6 F1 macro (using DeepLift) compared to 64.5 obtained by a full-text model. However, this does not apply to the other feature attribution methods ($\alpha$; $x\nabla x$; IG). 
To better understand this behavior, we conduct a correlation analysis between the importance rankings using any single feature attribution from (1) a model trained on the same domain with the evaluation data; and (2) a model trained on a different domain (out-of-domain trained model). High correlations suggest that if a feature attribution from an out-of-domain trained model produces similar importance distributions with that of an in-domain model, it will also lead to high predictive performance out-of-domain. Contrary to our initial assumption we found that the lower the correlation, the higher the predictive performance with FRESH. Results show low correlations when using $\alpha\nabla\alpha$ and DeepLift (highest FRESH performance). Surprisingly, IG and $x\nabla x$ (lowest FRESH performance) showed consistently strong correlations across all dataset pairs. Thus, we conclude that lower correlation scores indicate lower attachment to spurious correlations learned during training. We expand our discussion and show results for the correlation analysis in Appendix \ref{app:extended_correlation_analysis}.
% We generally note that any feature attribution performing well on the in-domain data, also performs well on the out-of-domain data.

Our findings therefore suggest that \textit{using FRESH in out-of-domain settings, can result to comparable performance with a model trained on full-text. However this highly depends on the choice of the feature attribution method}. 
% We also expect the in-domain performance of feature attribution approaches as rationale extractors for FRESH, to reflect how well they can perform in out-of-domain settings. 

\paragraph{HardKuma vs. FRESH:} We observe that HardKuma models are not significantly different compared to models trained on the full text in out-of-domain settings in more cases, when compared to FRESH (9 out of 12 and 5 out of 12 respectively). However, \emph{FRESH with $\alpha\nabla\alpha$ or DeepLift records higher predictive performance compared to HardKuma models (both in- and out-of-domain) in all cases}. We attribute this to the underlying model architectures, as FRESH uses \bert{} and HardKuma a bi-\texttt{LSTM}. As we discussed in \S\ref{par:select_then_predict}, we attempted using \bert{} for HardKuma models in the extractor and classifier similar to \citet{jain-etal-2020-learning}. However, the performance of HardKuma with \bert{} is at most comparable to when using a bi-\texttt{LSTM} similar to findings of \citet{guerreiro2021spectra}.

\subsection{Correlation between Post-hoc Explanation Faithfulness and FRESH Performance}

We hypothesize that a feature attribution with high scores for sufficiency and comprehensiveness, should extract rationales that result in high FRESH predictive performance. We expect that if our hypothesis is valid, faithfulness scores can serve as early indicators of FRESH performance, both on in-domain and out-of-domain settings.

Table \ref{tab:correlation_spearman_fresh_comp} shows the Spearman's ranking correlation ($\rho$) between FRESH F1 performance (see Table \ref{tab:F1_ece_select-then-predict-FRESH}) and comprehensiveness and sufficiency (see Table \ref{tab:faithfulness_feature_scoring_comp-aopc}). Correlation is computed using all feature scoring methods for each dataset pair. Results show that only 4 cases achieve statistically significant correlations (p-value $<$ 0.05) with only 3 out-of-domain and mostly between sufficiency and FRESH performance. We do not observe high correlations with comprehensiveness which is expected, as comprehensiveness evaluated the rationale's influence towards a model's prediction. Our findings refute our initial hypothesis and suggest that \emph{there is no clear correlation across all cases, between post-hoc explanation faithfulness and FRESH predictive performance. Therefore, sufficiency and comprehensiveness scores cannot be used as early indicators of FRESH predictive performance.}

\renewcommand{\arraystretch}{1.1}
\begin{table}[!t]
    \setlength\tabcolsep{2pt}
    \centering
    \small
    \begin{tabular}{l||lc|c}
     \textbf{Train} &   \textbf{Test}  &\multicolumn{2}{c}{$\boldsymbol\rho$}  \\
     & {\bf FRESH} & \textbf{Sufficiency} & \textbf{Comprehen.} \\\hline\hline
            \multirow{3}{*}{SST} &             SST &  \textbf{0.97} &           0.15 \\
         &        IMDB &           0.36 &           0.21 \\
         &        Yelp &   \textbf{0.90} &           0.56 \\ \hline
       \multirow{3}{*}{IMDB} &        IMDB &           0.69 &           0.87 \\
        &         SST &           0.65 &           0.23 \\
        &        Yelp &  \textbf{0.92} &  \textbf{0.92} \\ \hline
       \multirow{3}{*}{Yelp} &        Yelp &           0.82 &           0.55 \\
        &         SST &          -0.67 &          -0.67 \\
        &        IMDB &           0.87 &           0.56 \\ \hline
 \multirow{3}{*}{AmazDigiMu} &  AmazDigiMu &          -0.11 &           0.22 \\
  &   AmazInstr &           0.23 &           0.69 \\
  &  AmazPantry &           0.11 &           0.11 \\ \hline
 \multirow{3}{*}{AmazPantry} &  AmazPantry &           0.16 &           0.16 \\
  &  AmazDigiMu &           0.05 &           0.41 \\
  &   AmazInstr &           0.16 &           0.16 \\ \hline
  \multirow{3}{*}{AmazInstr} &   AmazInstr &           0.79 &           0.55 \\
   &  AmazDigiMu &           0.24 &           0.67 \\
   &  AmazPantry &           0.21 &            0.20 \\
    \end{tabular}
    \caption{Spearman's ranking correlation ($\rho$) between FRESH performance and comprehensiveness, sufficiency across all feature attribution approaches. \textbf{Bold} denotes statistically significant (p-value $\leq$ 0.05) correlations.}
    \label{tab:correlation_spearman_fresh_comp}
\end{table}

\section{Qualitative Analysis}

Table \ref{tab:qualitative} presents examples from a qualitative analysis we performed, aimed at better understanding out-of-domain post-hoc explanations. Rows with highlighted text in \colorbox{cyan}{blue} are from a model trained in the same domain as the presented example (ID), whilst those with highlighted text in \colorbox{red}{red} are from models trained on a different domain. Importance scores are computed using scaled attention ($\nabla \alpha \nabla$).

\renewcommand{\arraystretch}{1.3}
\begin{table*}[!t]
	\footnotesize
	\centering
	% {p{0.25\textwidth} | p{0.5\textwidth}}
	\begin{tabular}{ll|l}
		 \multicolumn{2}{c}{$\boldsymbol{\mathcal{M}}$ \textbf{Trained On}} & \textbf{Example} \\ \hline \hline
		\multirow{3}{*}{(1)}& AmazInstr (ID) &  \small{
			\colorbox{cyan!47.7}{\strut Work }\colorbox{cyan!80.0}{\strut great }\colorbox{cyan!64.0}{\strut and }\colorbox{cyan!23.2}{\strut sound }\colorbox{cyan!36.5}{\strut good }
		}\\
		& AmazDigiMu  & \small{
			\colorbox{red!0.0}{\strut Work }\colorbox{red!100.0}{\strut great }\colorbox{red!55.3}{\strut and }\colorbox{red!32.1}{\strut sound }\colorbox{red!35.5}{\strut good }
		}\\
		& AmazPantry & \small{
			\colorbox{red!22.6}{\strut Work }\colorbox{red!100.0}{\strut great }\colorbox{red!25.1}{\strut and }\colorbox{red!0.0}{\strut sound }\colorbox{red!6.0}{\strut good }
		} \\ \hline \hline

		\multirow{3}{*}{(2)}&AmazPantry (ID) & \colorbox{cyan!100.0}{\strut Delicious }\colorbox{cyan!57.8}{\strut and }\colorbox{cyan!23.3}{\strut at }\colorbox{cyan!15.6}{\strut a }\colorbox{cyan!32.0}{\strut good }\colorbox{cyan!8.1}{\strut price }\colorbox{cyan!4.9}{\strut . }\colorbox{cyan!33.6}{\strut would }\colorbox{cyan!43.6}{\strut recommend }\colorbox{cyan!4.9}{\strut . }
		 \\ 
		& AmazDigiMu & \colorbox{red!19.8}{\strut Delicious }\colorbox{red!60.2}{\strut and }\colorbox{red!69.4}{\strut at }\colorbox{red!44.3}{\strut a }\colorbox{red!59.4}{\strut good }\colorbox{red!52.5}{\strut price }\colorbox{red!0.0}{\strut . }\colorbox{red!75.7}{\strut would }\colorbox{red!100.0}{\strut recommend }\colorbox{red!82.1}{\strut . }
			\\
		& AmazInstr & \colorbox{red!34.3}{\strut Delicious }\colorbox{red!100.0}{\strut and }\colorbox{red!76.2}{\strut at }\colorbox{red!65.5}{\strut a }\colorbox{red!79.9}{\strut good }\colorbox{red!0.0}{\strut price }\colorbox{red!53.6}{\strut . }\colorbox{red!59.1}{\strut would }\colorbox{red!74.1}{\strut recommend }\colorbox{red!0.8}{\strut . }
		 \\ \hline\hline
		
		\multirow{3}{*}{(3)}& SST (ID) &  \colorbox{cyan!44.8}{\strut A }\colorbox{cyan!74.6}{\strut painfully }\colorbox{cyan!100.0}{\strut funny }\colorbox{cyan!8.0}{\strut ode }\colorbox{cyan!2.6}{\strut to }\colorbox{cyan!3.8}{\strut bad }\colorbox{cyan!6.2}{\strut behavior } \\
	    & IMDB &  
		\colorbox{red!90.7}{\strut A }\colorbox{red!0.0}{\strut painfully }\colorbox{red!51.4}{\strut funny }\colorbox{red!100.0}{\strut ode }\colorbox{red!15.6}{\strut to }\colorbox{red!21.7}{\strut bad }\colorbox{red!27.5}{\strut behavior } \\
		& Yelp & \colorbox{red!0.0}{\strut A }\colorbox{red!16.3}{\strut painfully }\colorbox{red!0.6}{\strut funny }\colorbox{red!41.2}{\strut ode }\colorbox{red!82.4}{\strut to }\colorbox{red!48.9}{\strut bad }\colorbox{red!100.0}{\strut behavior }\\ \hline \hline
		
		\multirow{3}{*}{(4)}& Yelp (ID) & \colorbox{cyan!3.5}{\strut The }\colorbox{cyan!16.8}{\strut kouign }\colorbox{cyan!43.5}{\strut - }\colorbox{cyan!37.3}{\strut amann }\colorbox{cyan!84.5}{\strut is }\colorbox{cyan!83.0}{\strut so }\colorbox{cyan!83.9}{\strut amazing }\colorbox{cyan!45.2}{\strut ... }\colorbox{cyan!78.8}{\strut must }\colorbox{cyan!82.1}{\strut taste }\colorbox{cyan!78.8}{\strut to }\colorbox{cyan!75.2}{\strut appreciate }\colorbox{cyan!58.3}{\strut . } \\
		& SST & \colorbox{red!40.2}{\strut The }\colorbox{red!18.1}{\strut kouign }\colorbox{red!5.6}{\strut - }\colorbox{red!1.8}{\strut amann }\colorbox{red!99.0}{\strut is }\colorbox{red!82.5}{\strut so }\colorbox{red!89.3}{\strut amazing }\colorbox{red!33.3}{\strut ... }\colorbox{red!22.9}{\strut must }\colorbox{red!52.0}{\strut taste }\colorbox{red!26.0}{\strut to }\colorbox{red!45.9}{\strut appreciate }\colorbox{red!66.1}{\strut . } \\
		& IMDB & \colorbox{red!32.7}{\strut The }\colorbox{red!20.2}{\strut kouign }\colorbox{red!13.8}{\strut - }\colorbox{red!28.7}{\strut amann }\colorbox{red!92.7}{\strut is }\colorbox{red!100.0}{\strut so }\colorbox{red!76.1}{\strut amazing }\colorbox{red!17.4}{\strut ... }\colorbox{red!85.5}{\strut must }\colorbox{red!26.6}{\strut taste }\colorbox{red!27.3}{\strut to }\colorbox{red!25.8}{\strut appreciate }\colorbox{red!39.7}{\strut . } \\ 
			
	\end{tabular}
	\caption{True examples of highlights with $\alpha\nabla\alpha$ using a model trained on data from the same distribution as the example (ID; with \colorbox{cyan!80}{blue highlights}) and two models trained on a different dataset (with \colorbox{red}{red highlights}).}
	\label{tab:qualitative}
\end{table*}

In Example (1), we observe that models trained on two closely related tasks (AmazInstr and AmazDigiMu) place more importance to the phrase ``sound good''. On the contrary, the model trained on AmazPantry which has not encountered such phrases during training, mostly focuses on ``Work great''. This is expected as the term ``sound'' is not typical of pantry reviews. Similarly, in Example (2) from the AmazPantry dataset, the in-domain model focuses on a domain-specific word ``delicious''. On the contrary, the two models trained on music-related tasks focus on more generic terms such as ``good'' and ``would recommend''. 
In Example (3) the model trained on Yelp focuses mostly on the word ``behavior'', a term we consider more relevant to restaurant reviews rather than movie reviews. In comparison, the other models which are both trained on movie reviews focus both on the term ``funny''. In Example (4), again the two movie-review models focus on more generic terms (i.e. ``amazing'') compared to ``must taste'' that the model trained in-domain (i.e. Yelp) identifies as important.

Overall, results show that rationales from models applied to a different domain (other than that they were trained for), comprise of terms that are mostly present within the domain they were trained for. This can partly explain the performance of out-of-domain FRESH classifiers. Similar to \citep{adebayo-nips}, our assumption is that a model's inability to generalize to other domains, is based on the model latching on to specific features from the training dataset.

\section{Conclusion}

We conducted an extensive empirical study to assess the faithfulness of post-hoc explanations (i.e. using feature attribution approaches) and performance of select-then-predict (i.e. inherently faithful) models in out-of-domain settings.
Our findings highlight, that using sufficiency and comprehensiveness to evaluate post-hoc explanation faithfulness out-of-domain can be misleading. 
% Our findings highlight the implications of using sufficiency and comprehensiveness to evaluate explanation faithfulness for standalone feature attributions. 
%Similar to \citet{madsen2021evaluating}, our work 
To address this issue, we suggest comparing faithfulness of post-hoc explanations to a random attribution baseline for a more robust evaluation.
We also show that select-then-predict models, which are inherently faithful, perform surprisingly well in out-of-domain settings. Despite performance degradation, in many cases their performance is comparable to those of full-text trained models. %HardKuma models perform in more cases comparably to their full-text counterparts in out-of-domain compared to FRESH. However, FRESH models record higher predictive performance, as they use \bert{}.
% Finally, we also show that $\alpha\nabla\alpha$ is largely the best performing feature attribution in both sufficiency, comprehensiveness (similar to findings of \citet{serrano-smith-2019-attention}) and FRESH. 
% As such, we argue that it should be evaluated further for its faithfulness and plausibility. 
In future work, we aim to explore methods for improving the evaluation of faithfulness for out-of-domain post-hoc explanations.

\section*{Acknowledgements}

We would like to thank Katerina Margatina and Tulika Bose for their insightful feedback in a preliminary version of this paper. GC and NA are supported by EPSRC grant EP/V055712/1, part of the European Commission CHIST-ERA programme, call 2019 XAI: Explainable Machine Learning-based Artificial Intelligence.

\bibliographystyle{acl_natbib}
\bibliography{biblio}

\appendix
\newpage

\section{Dataset Characteristics
\label{app:extended_data}}

Table \ref{tab:extednded_data_characteristics} presents extended data characteristics for all datasets. We present information across the three data splits, including: (1) The average sequence length; (2) The number of documents in each split and (3) the number of documents under each label. 

Our dataset selection was highly motivated for also examining the differences when we have gradual shifts in-domain. For example for the triplet SST - IMDB - YELP, two datasets are closely associated (SST, IMDB) as they are movie reviews, whilst Yelp is a task for classifying restaurant reviews. Similarly, AmazDigiMu and AmazInstr share similar characteristics, as they are reviews about items related to music. On the contrary, AmazPantry consists of reviews about pantry items. This is also the primary reason why we focused on text classification tasks, as it is easier to control for the output and other parameters, whilst allowing for control over the task it-self. 

\setlength\tabcolsep{2pt}
\begin{table}[!b]
    \centering
    \small
    \begin{tabular}{ll||ccc}
    Dataset &                       &   Train &    Dev &   Test \\ \hline \hline
    \multirow{4}{*}{SST} &  Avg. Seq. Length &      17 &     17 &     17 \\
         &      No. of documents&    6,920 &    872 &   1,821 \\
         &       Docs in label-0 &    3,310 &    428 &    912 \\
         &       Docs in label-1 &    3,610 &    444 &    909 \\ \hline
       \multirow{4}{*}{IMDB} &  Avg. Seq. Length &     241 &    248 &    247 \\
        &      No. of documents &   20,000 &   2,500 &   2,500 \\
        &       Docs in label-0 &    9,952 &   1,275 &   1,273 \\
        &       Docs in label-1 &   10,048 &   1,225 &   1,227 \\ \hline
       \multirow{4}{*}{Yelp} &  Avg. Seq. Length &     154 &    154 &    153 \\
        &      No. of documents&  476,000 &  84,000 &  38,000 \\
        &       Docs in label-0 &  238,000 &  42,000 &  19,000 \\
        &       Docs in label-1 &  238,000 &  42,000 &  19,000 \\\hline
 \multirow{4}{*}{AmazDigiMu} & Avg. Seq. Length &      38 &     39 &     38 \\
  &      No. of documents&  122,552 &  21,627 &  25,444 \\
  &       Docs in label-0 &    2,893 &    510 &    601 \\
  &       Docs in label-1 &    4,907 &    866 &   1,019 \\
  &       Docs in label-2 &  114,752 &  20,251 &  23,824 \\\hline
 \multirow{4}{*}{AmazPantry} &  Avg. Seq. Length &      24 &     24 &     24 \\
  &      No. of documents&   99,423 &  17,546 &  20,642 \\
  &       Docs in label-0 &    4,995 &    881 &   1,037 \\
  &       Docs in label-1 &    6,579 &   1,161 &   1,366 \\
  &       Docs in label-2 &   87,849 &  15,504 &  18,239 \\ \hline
  \multirow{4}{*}{AmazInstr} &  Avg. Seq. Length &      66 &     66 &     65 \\
   &      No. of documents&  167,145 &  29,497 &  34,702 \\
   &       Docs in label-0 &   10,651 &   1,879 &   2,211 \\
   &       Docs in label-1 &   11,581 &   2,044 &   2,404 \\
   &       Docs in label-2 &  144,913 &  25,574 &  30,087 \\ 
\end{tabular}
    \caption{Extended dataset characteristics}
    \label{tab:extednded_data_characteristics}
\end{table}

\section{Models and Hyper-parameters
\label{app:extended_models}}

\paragraph{For feature attributions:} We use \bert{}-base with pre-trained weights from the Huggingface library \citep{Wolf2019HuggingFacesTS}. We use the AdamW optimizer \citep{loshchilov2017decoupled} with an initial learning rate of $1e-5$ for fine-tuning \bert{} and $1e-4$ for the fully-connected classification layer. We train our models for 3 epochs using a linear scheduler with 10\% of the data in the first epoch as warm-up. We also use a grad-norm of 1 and select the model with the lowest loss on the development set. All models are trained across 5 random seeds and we report the average and standard deviation. We present their test-set performance in Table \ref{tab:F1_ece} and their development set performance in Table \ref{tab:f1-macro_dev}.

\paragraph{For FRESH:} For the rationale extractor, we use the same model for extracting rationales from feature attributions. For the classifier (trained only on the extracted rationales), we also use \bert{}-base with the same optimizer configuration and scheduler warm-up steps. We also use a grad-norm of 1 and select the model with the lowest loss on the development set. We train across 5 random seeds for 5 epochs. 

In Table \ref{tab:F1_ece} we present full-text \bert{}-base F1-macro scores averaged across 5 random seeds with standard deviations included in the brackets. Additionally, we present the mean Expected Calibration Error (ECE) scores. Finally, in Table \ref{tab:f1-macro_dev} we present the in-domain F1-macro performance and loss on the development set. 

\setlength\tabcolsep{4pt}
\begin{table}[!t]
    \centering
    \small
    \begin{tabular}{l||ccc}
        \textbf{Trained On} & \textbf{Tested On} & \textbf{F1} & \textbf{ECE}  \\ \hline \hline
        \multirow{3}{*}{SST} &   SST &  90.1 (0.3) &  4.4 (0.7) \\
        &        IMDB &  84.3 (0.6) &  7.1 (0.6) \\
        &        Yelp &  87.9 (2.3) &  4.2 (2.3) \\ \hline
        \multirow{3}{*}{IMDB}  & IMDB &  91.1 (0.4) &  4.7 (0.6) \\
       &         SST &  85.8 (2.0) &  5.8 (0.8) \\
       &        Yelp &  91.0 (1.2) &  0.9 (0.2) \\ \hline
         \multirow{3}{*}{Yelp} &  Yelp &  96.9 (0.1) &  2.2 (0.1) \\
       &         SST &  86.8 (1.7) &  8.5 (0.9) \\
       &        IMDB &  88.6 (0.3) &  7.9 (0.6) \\ \hline
        \multirow{3}{*}{AmazDigiMu} & AmazDigiMu &  70.6 (0.9) &  2.3 (0.1) \\
 &   AmazInstr &  61.2 (1.8) &  5.4 (0.2) \\
 &  AmazPantry &  64.6 (1.0) &  4.3 (0.4) \\ \hline
        \multirow{3}{*}{AmazPantry} & AmazPantry &  70.2 (1.1) &  3.8 (0.4) \\
  &  AmazDigiMu &  59.5 (0.7) &  3.2 (0.5) \\
  &   AmazInstr &  64.5 (2.6) &  4.9 (0.9) \\ \hline
        \multirow{3}{*}{AmazInstr} & AmazInstr &  71.5 (0.4) &  3.9 (0.5) \\
   &  AmazDigiMu &  61.3 (0.3) &  3.2 (0.2) \\
   &  AmazPantry &  68.2 (0.7) &  4.1 (0.5) \\
    \end{tabular}
    \caption{F1 macro performance and Expected Calibration Error (ECE) (five runs) with standard deviation, of full-text \bert{}-base models. }
    \label{tab:F1_ece}
\end{table}

\begin{table}[!t]
    \centering
    \small
    \begin{tabular}{l|cc}
         Dataset &          F1 &  Dev. Loss \\ \hline
        SST &  89.9 (0.3) &  2.4 (0.0) \\
       IMDB &  92.0 (0.3) &  1.8 (0.0) \\
       Yelp &  96.8 (0.1) &  0.9 (0.0) \\
     AmazDigiMu &  67.6 (1.1) &  1.3 (0.0) \\
     AmazPantry &  69.5 (1.4) &  1.9 (0.1) \\
      AmazInstr &  72.1 (0.5) &  1.9 (0.1) \\
    \end{tabular}
    \caption{F1-macro predictive performance (five runs) with standard deviation,  of \bert{}-base models trained on the full text. We also include the development loss.}
    \label{tab:f1-macro_dev}
\end{table}

\paragraph{For HardKuma: } We use the 300-dimensional pre-trained GloVe embeddings from the 840B release \citep{pennington-etal-2014-glove} as word representations and keep them frozen during training. The rationale extractor (which generates the rationale mask $z$) is a 200-d bi-directional \texttt{LSTM} layer (bi-\texttt{LSTM}) \citep{hochreiter1997long} similar to \citep{bastings-etal-2019-interpretable, guerreiro2021spectra}. We use the Adam optimizer \citep{kingma2014adam} for all models with a learning rate between $1e-5$ and $1e-4$ and a weight decay of $1e-5$. We also enforce a grad-norm of 5 and train for 20 epochs across 5 random seeds. Similar to \citet{guerreiro2021spectra} we select the model with the highest F1-macro score on the development set and find that tuning the  Lagrangian relaxation algorithm parameters beneficial to model predictive performance. 
We also attempted training HardKuma models with \bert{}-base, similar to \citet{jain-etal-2020-learning}, however we found performance to be at best comparable with our \texttt{LSTM} variant, as in \citet{guerreiro2021spectra}, even after hyperparameter tuning. 

\setlength\tabcolsep{4pt}
\begin{table}[!t]
    \centering
    \small
    \begin{tabular}{l||ccc}
        \textbf{Trained On} & \textbf{Tested On} & \textbf{F1} & \textbf{ECE}  \\ \hline \hline
        \multirow{3}{*}{SST} &   SST &  81.7 (0.9) &   3.2 (0.7) \\
         &        IMDB &  71.9 (0.9) &   4.9 (2.8) \\
         &        Yelp &  68.7 (3.2) &   5.8 (5.1) \\ \hline
        \multirow{3}{*}{IMDB}  &IMDB &  87.4 (0.9) &   4.7 (1.8) \\
        &         SST &  77.5 (2.0) &   6.2 (1.4) \\
        &        Yelp &  41.0 (5.3) &  39.4 (7.3) \\\hline
         \multirow{3}{*}{Yelp} & Yelp &  96.0 (0.0) &   0.5 (0.2) \\
        &         SST &  80.4 (0.8) &   1.9 (0.7) \\
        &        IMDB &  84.5 (1.0) &   5.0 (1.3) \\\hline
        \multirow{3}{*}{AmazDigiMu} & AmazDigiMu &  67.6 (0.3) &   0.5 (0.1) \\
  &   AmazInstr &  54.2 (1.1) &   2.6 (0.6) \\
  &  AmazPantry &  55.3 (0.4) &   1.9 (0.5) \\ \hline
        \multirow{3}{*}{AmazPantry} & AmazPantry &  67.9 (0.4) &   0.7 (0.4) \\
  &  AmazDigiMu &  50.9 (1.9) &   1.9 (0.6) \\
  &   AmazInstr &  55.9 (2.2) &   2.8 (0.9) \\ \hline
        \multirow{3}{*}{AmazInstr} & AmazInstr &  67.2 (0.7) &   1.2 (0.4) \\
   &  AmazDigiMu &  54.3 (1.4) &   1.1 (0.1) \\
   &  AmazPantry &  61.1 (1.5) &   1.5 (0.6) \\
    \end{tabular}
    \caption{F1 macro performance and Expected Calibration Error (ECE) of a full-text \texttt{LSTM} classifier trained on an in-domain dataset and tested on their in-domain test-set and two other out-of-domain datasets. }
    \label{tab:F1_ece_HARDKUMA}
\end{table}

\noindent All experiments are run on a single NVIDIA Tesla V100 GPU.

\renewcommand{\arraystretch}{1.2}
\begin{table*}[!t]
    \setlength\tabcolsep{1pt}
    \centering
    % \scriptsize
    \small
    \begin{tabular}{l||cc|ccc}

        \textbf{Train} & \textbf{Test}  & \textbf{Full-text} & \multicolumn{3}{c}{\textbf{HardKuma}}  \\
         &   & \textbf{F1} & \textbf{F1} & \textbf{ECE} &\textbf{L (\%)} \\ \hline \hline
          \multirow{3}{*}{SST} &         SST &    81.7 (0.9) &           77.6 (1.4) &   3.8 (0.8) &  56.8 (26.2) \\
                             &        IMDB &    71.9 (0.9) &  \textbf{65.7}(15.1) &   7.4 (6.4) &  39.5 (33.5) \\
                             &        Yelp &    68.7 (3.2) &  \textbf{67.7}(11.6) &   9.9 (4.4) &  32.7 (30.7) \\ \hline
       \multirow{3}{*}{IMDB} &        IMDB &    87.4 (0.9) &           82.0 (0.6) &   3.5 (1.6) &    1.9 (0.2) \\
                             &         SST &    77.5 (2.0) &           73.6 (2.2) &   7.3 (5.3) &   16.8 (2.7) \\
                             &        Yelp &    41.0 (5.3) &   \textbf{47.2}(5.8) &  24.7 (6.3) &    3.1 (2.0) \\  \hline
       \multirow{3}{*}{Yelp} &        Yelp &    96.0 (0.0) &           92.4 (0.3) &   3.0 (0.7) &    7.4 (0.7) \\
                             &         SST &    80.4 (0.8) &           72.4 (0.8) &  10.9 (0.8) &   14.1 (1.2) \\
                             &        IMDB &    84.5 (1.0) &           73.3 (3.5) &  19.1 (3.8) &    4.7 (0.7) \\  \hline
 \multirow{3}{*}{AmazDigiMu} &  AmazDigiMu &    67.6 (0.3) &           66.8 (0.5) &   0.7 (0.5) &   18.4 (0.5) \\
                             &   AmazInstr &    54.2 (1.1) &   \textbf{53.3}(1.2) &   4.1 (2.0) &   25.8 (6.1) \\
                             &  AmazPantry &    55.3 (0.4) &   \textbf{54.7}(1.4) &   3.6 (1.4) &   27.8 (3.6) \\  \hline
 \multirow{3}{*}{AmazPantry} &  AmazPantry &    67.9 (0.4) &           66.6 (0.5) &   1.3 (0.4) &   18.9 (1.1) \\
                             &  AmazDigiMu &    50.9 (1.9) &   \textbf{51.0}(0.6) &   1.9 (0.6) &   11.2 (3.3) \\
                             &   AmazInstr &    55.9 (2.2) &   \textbf{57.4}(1.2) &   2.8 (0.6) &   18.2 (1.3) \\  \hline
  \multirow{3}{*}{AmazInstr} &   AmazInstr &    67.2 (0.7) &   \textbf{66.7}(0.8) &   1.9 (0.6) &   19.2 (1.5) \\
                             &  AmazDigiMu &    54.3 (1.4) &   \textbf{53.7}(1.2) &   1.9 (0.4) &   13.9 (2.9) \\
                             &  AmazPantry &    61.1 (1.5) &   \textbf{59.5}(1.4) &   2.8 (0.5) &   24.4 (2.8) \\
    \end{tabular}
    \caption{F1 macro performance (five runs) with standard deviation for HardKuma models and the selected rationale length (L).  \textbf{Bold} denotes no significant difference between HardKuma and Full-text (t-test; p $>$ 0.05).}
    \label{tab:F1_ece_hardkuma_STD}
\end{table*}

\setlength\tabcolsep{2pt}
\begin{table*}[!t]
    \centering
    \scriptsize
    \begin{tabular}{l||cc||ccccc || ccccc}
        \textbf{Train} & \textbf{Test} & \textbf{Full-Text} & \multicolumn{5}{c||}{\textbf{F1}} & \multicolumn{5}{c}{\textbf{ECE}} \\
        & & & $\alpha\nabla\alpha$ & $\alpha$ & DeepLift  & $x\nabla x$ & IG & $\alpha\nabla\alpha$ & $\alpha$ & DeepLift  & $x\nabla x$ & IG \\ \hline \hline
         \multirow{3}{*}{SST (20\%)} &         SST &   90.1 (0.3) &          87.7 (0.4) &          81.1 (1.0) &          84.4 (0.7) &  76.3 (0.5) &  76.8 (0.3) &            7.6 (1.6) &     6.0 (0.7) &    7.5 (0.5) &     2.7 (1.2) &          2.8 (1.3) \\
                             &        IMDB &   84.3 (0.6) &          81.8 (0.2) &          52.6 (2.1) &          64.0 (2.1) &  55.0 (1.7) &  56.3 (0.4) &           14.2 (1.2) &    21.1 (4.0) &   21.3 (3.5) &    18.2 (1.3) &         21.1 (0.7) \\
                             &        Yelp &   87.9 (2.3) &  \textbf{88.1}(0.0) &          72.6 (4.0) &          75.4 (2.3) &  59.6 (3.8) &  63.9 (1.1) &            8.1 (1.5) &     7.8 (3.2) &   11.5 (1.5) &     7.8 (4.3) &   7.8 (2.3) \\ \hline 
       \multirow{3}{*}{IMDB  (2\%)} &        IMDB &   91.1 (0.4) &          87.9 (0.2) &          80.4 (0.9) &          87.2 (0.4) &  59.8 (0.2) &  59.7 (0.6) &            8.2 (0.1) &     5.6 (1.5) &    7.7 (0.5) &     5.9 (3.2) &          5.9 (2.4) \\
                             &         SST &   85.8 (2.0) &          80.9 (0.5) &          71.8 (1.0) &          70.1 (0.5) &  69.6 (0.5) &  70.7 (1.7) &           13.1 (0.3) &     9.2 (1.9) &   22.6 (1.6) &     7.2 (1.0) &          5.9 (1.3) \\
                             &        Yelp &   91.0 (1.2) &          87.8 (0.1) &          82.0 (0.2) &          79.4 (1.4) &  69.0 (0.6) &  69.1 (0.4) &            7.3 (0.5) &     2.0 (1.9) &   14.6 (1.8) &     6.5 (1.4) &   6.8 (0.3) \\ \hline 
       \multirow{3}{*}{Yelp  (10\%)} &        Yelp &   96.9 (0.1) &          94.0 (0.0) &          90.4 (0.2) &          93.6 (0.3) &  70.5 (0.2) &  71.9 (0.1) &            4.3 (0.4) &     5.5 (0.4) &    3.6 (0.3) &     1.7 (0.8) &          2.2 (0.4) \\
                             &         SST &   86.8 (1.7) &          59.3 (0.6) &          69.8 (1.1) &          67.2 (1.5) &  67.7 (0.5) &  69.3 (0.8) &           33.5 (1.3) &    22.6 (0.8) &   28.8 (0.3) &     9.9 (0.4) &         10.8 (0.2) \\
                             &        IMDB &   88.6 (0.3) &          78.0 (0.4) &          64.5 (0.3) &          66.6 (0.5) &  53.0 (0.4) &  55.8 (0.1) &           17.4 (0.9) &    22.5 (1.4) &   29.8 (1.4) &    17.9 (1.7) &  18.1 (0.2)  \\ \hline
 \multirow{3}{*}{AmazDigiMu (20\%)} &  AmazDigiMu &   70.6 (0.9) &          66.1 (1.8) &          63.4 (1.0) &  \textbf{65.8}(2.6) &  51.9 (2.0) &  65.8 (2.6) &            2.8 (0.4) &     2.2 (0.9) &    2.7 (0.7) &     2.4 (0.9) &          2.7 (0.7) \\
                             &   AmazInstr &   61.2 (1.8) &  \textbf{58.0}(0.8) &  \textbf{57.2}(1.2) &  \textbf{57.4}(1.2) &  46.0 (0.6) &  57.2 (1.2) &            8.2 (1.0) &     6.7 (1.5) &    8.3 (1.3) &     6.3 (1.8) &          6.7 (1.5) \\
                             &  AmazPantry &   64.6 (1.0) &          59.1 (0.3) &          56.5 (1.2) &          56.5 (1.7) &  44.8 (0.8) &  44.8 (0.8) &            6.5 (0.8) &     5.6 (1.4) &    7.1 (1.6) &     5.8 (1.6) &   5.8 (1.6) \\ \hline 
 \multirow{3}{*}{AmazPantry (20\%)} &  AmazPantry &   70.2 (1.1) &          67.3 (0.5) &          62.6 (1.0) &          67.2 (0.0) &  48.6 (1.7) &  48.7 (2.7) &            4.9 (0.3) &     3.8 (0.3) &    4.9 (0.3) &     4.1 (1.0) &          4.3 (1.3) \\
                             &  AmazDigiMu &   59.5 (0.7) &  \textbf{57.7}(0.6) &          54.6 (0.9) &          56.2 (0.0) &  41.2 (0.4) &  57.7 (0.6) &            3.6 (0.4) &     2.7 (0.2) &    3.7 (0.1) &     1.8 (0.9) &          3.6 (0.4) \\
                             &   AmazInstr &   64.5 (2.6) &  \textbf{63.8}(0.4) &          58.0 (1.9) &  \textbf{63.6}(0.2) &  40.1 (1.1) &  40.3 (2.5) &            6.6 (0.4) &     5.3 (0.7) &    6.5 (0.4) &     5.7 (1.5) &   5.8 (1.9)  \\ \hline
  \multirow{3}{*}{AmazInstr (20\%)} &   AmazInstr &   71.5 (0.4) &          69.8 (0.3) &          62.1 (2.3) &          69.7 (0.3) &  45.6 (4.7) &  48.6 (2.7) &            5.6 (0.5) &     3.6 (0.7) &    5.9 (0.3) &     2.4 (1.0) &          3.2 (1.1) \\
                             &  AmazDigiMu &   61.3 (0.3) &  \textbf{60.0}(0.7) &          53.2 (1.7) &          57.8 (0.4) &  43.8 (3.3) &  60.0 (0.7) &            3.5 (0.4) &     1.8 (0.3) &    4.1 (0.2) &     1.4 (0.1) &          3.5 (0.4) \\
                             &  AmazPantry &   68.2 (0.7) &          64.5 (0.7) &          56.3 (1.9) &          63.1 (0.3) &  44.6 (3.9) &  47.6 (2.6) &            5.7 (0.4) &     4.0 (0.3) &    6.0 (0.3) &     2.7 (1.2) &   3.6 (0.9) \\
    \end{tabular}
    \caption{F1 macro performance of FRESH models (TopK rationales) with standard deviation in brackets and Expected Calibration Error (ECE) scores. For reference we include the in-domain performance of full-text models. \textbf{Bold} denotes no significant difference between FRESH and Full-text (t-test; p $>$ 0.05)}
    \label{tab:F1_ece_select-then-predict-FRESH_TOPK}
\end{table*}

\setlength\tabcolsep{2pt}
\begin{table*}[!t]
    \centering
    \scriptsize
    \begin{tabular}{l||cc||ccccc || ccccc}
        \textbf{Train} & \textbf{Test} & \textbf{Full-Text} & \multicolumn{5}{c||}{\textbf{F1}} & \multicolumn{5}{c}{\textbf{ECE}} \\
        & & & $\alpha\nabla\alpha$ & $\alpha$ & DeepLift  & $x\nabla x$ & IG & $\alpha\nabla\alpha$ & $\alpha$ & DeepLift  & $x\nabla x$ & IG \\ \hline \hline
         \multirow{3}{*}{SST (20\%)} &         SST &   90.1 (0.3) &          87.1 (0.8) &  80.7 (0.4) &  79.7 (1.5) &  77.8 (0.6) &  79.7 (1.5) &            5.9 (0.5) &     4.2 (1.9) &    5.8 (2.0) &     2.5 (0.9) &          5.8 (2.0) \\
                             &        IMDB &   84.3 (0.6) &          80.3 (0.5) &  58.8 (0.4) &  64.9 (1.5) &  53.1 (0.7) &  64.9 (1.5) &           13.3 (0.6) &    19.7 (2.8) &   15.3 (1.7) &    19.0 (2.6) &         15.3 (1.7) \\
                             &        Yelp &   87.9 (2.3) &  \textbf{88.1}(0.3) &  74.8 (1.0) &  69.5 (0.9) &  71.7 (1.1) &  88.1 (0.3) &            5.4 (0.3) &     4.0 (2.7) &    9.4 (3.1) &     3.1 (1.8) &   5.4 (0.3)  \\\hline
       \multirow{3}{*}{IMDB (2\%)} &        IMDB &   91.1 (0.4) &          83.2 (0.1) &  75.6 (0.6) &  82.5 (0.8) &  62.7 (0.2) &  82.5 (0.8) &            7.1 (1.4) &     4.8 (1.4) &    7.6 (1.5) &     3.8 (1.3) &          7.6 (1.5) \\
                             &         SST &   85.8 (2.0) &          80.1 (1.1) &  74.7 (1.2) &  66.7 (0.6) &  71.6 (1.2) &  80.1 (1.1) &            8.1 (0.9) &     3.1 (1.4) &   20.1 (1.7) &     4.2 (0.7) &          8.1 (0.9) \\
                             &        Yelp &   91.0 (1.2) &          87.0 (0.3) &  80.8 (1.3) &  69.2 (4.4) &  73.8 (0.8) &  87.0 (0.3) &            3.4 (2.0) &     2.8 (0.2) &   15.8 (2.1) &     8.1 (1.4) &   3.4 (2.0) \\\hline 
       \multirow{3}{*}{Yelp (10\%)} &        Yelp &   96.9 (0.1) &          91.8 (0.5) &  81.7 (0.3) &  89.0 (0.7) &  81.8 (0.2) &  89.0 (0.7) &            5.4 (0.4) &     3.7 (0.9) &    5.3 (0.4) &     4.0 (0.7) &          5.3 (0.4) \\
                             &         SST &   86.8 (1.7) &          65.5 (2.2) &  71.3 (1.3) &  68.4 (1.0) &  68.7 (0.5) &  65.5 (2.2) &           26.6 (2.0) &    15.3 (2.8) &   23.7 (2.4) &     9.0 (0.7) &         26.6 (2.0) \\
                             &        IMDB &   88.6 (0.3) &          75.3 (1.2) &  62.1 (0.9) &  67.5 (0.2) &  55.8 (0.4) &  67.5 (0.2) &           19.2 (0.7) &    15.1 (0.6) &   24.3 (1.6) &    17.6 (0.7) &  24.3 (1.6)  \\ \hline
 \multirow{3}{*}{AmazDigiMu (20\%)} &  AmazDigiMu &   70.6 (0.9) &          65.8 (1.5) &  60.1 (2.3) &  59.5 (4.0) &  55.9 (2.4) &  59.5 (4.0) &            2.8 (0.4) &     2.4 (1.0) &    3.2 (0.4) &     2.6 (1.1) &          3.2 (0.4) \\
                             &   AmazInstr &   61.2 (1.8) &          57.0 (0.9) &  51.8 (2.0) &  50.8 (1.8) &  47.5 (0.6) &  51.8 (2.0) &            8.2 (1.0) &     6.6 (2.1) &    8.5 (1.0) &     6.4 (2.1) &          6.6 (2.1) \\
                             &  AmazPantry &   64.6 (1.0) &          57.7 (0.6) &  51.6 (2.0) &  51.4 (2.6) &  47.5 (1.2) &  47.5 (1.2) &            6.7 (0.8) &     5.7 (1.8) &    7.5 (0.5) &     6.1 (1.8) &   6.1 (1.8)  \\\hline
 \multirow{3}{*}{AmazPantry (20\%)} &  AmazPantry &   70.2 (1.1) &  \textbf{63.5}(3.6) &  62.0 (0.4) &  58.0 (1.0) &  50.0 (2.1) &  58.0 (1.0) &            4.4 (0.4) &     3.8 (0.6) &    5.0 (0.9) &     4.3 (0.9) &          5.0 (0.9) \\
                             &  AmazDigiMu &   59.5 (0.7) &  \textbf{53.7}(3.6) &  52.0 (1.4) &  46.7 (0.7) &  44.4 (2.7) &  53.7 (3.6) &            3.2 (0.2) &     2.8 (0.5) &    2.8 (0.9) &     1.9 (0.7) &          3.2 (0.2) \\
                             &   AmazInstr &   64.5 (2.6) &  \textbf{59.1}(3.9) &  56.1 (1.5) &  51.4 (0.6) &  42.6 (3.6) &  56.1 (1.5) &            5.8 (0.4) &     5.7 (1.0) &    5.7 (1.5) &     5.7 (1.5) &   5.7 (1.0) \\ \hline
  \multirow{3}{*}{AmazInstr (20\%)} &   AmazInstr &   71.5 (0.4) &          66.3 (1.1) &  52.2 (2.3) &  60.9 (0.8) &  53.4 (1.2) &  60.9 (0.8) &            4.6 (0.2) &     4.2 (0.6) &    5.2 (0.9) &     3.7 (1.4) &          5.2 (0.9) \\
                             &  AmazDigiMu &   61.3 (0.3) &          56.5 (0.6) &  47.0 (1.4) &  52.1 (0.3) &  48.3 (1.2) &  56.5 (0.6) &            2.9 (0.2) &     1.9 (0.4) &    3.3 (0.6) &     2.0 (0.6) &          2.9 (0.2) \\
                             &  AmazPantry &   68.2 (0.7) &          62.4 (0.9) &  49.2 (1.7) &  57.4 (0.6) &  51.0 (1.3) &  51.0 (1.3) &            4.6 (0.3) &     4.6 (0.5) &    5.2 (0.8) &     4.5 (0.8) &   4.5 (0.8) \\
    \end{tabular}
    \caption{F1 macro performance of FRESH models (Contiguous rationales) with standard deviation in brackets and Expected Calibration Error (ECE) scores. For reference we include the in-domain performance of full-text models. \textbf{Bold} denotes no significant difference between FRESH and Full-text (t-test; p $>$ 0.05)}
    \label{tab:F1_ece_select-then-predict-FRESH_Contiguous}
\end{table*}

\section{HardKuma - Extended
\label{app:exteded_hardkuma}}

In Table \ref{tab:F1_ece_HARDKUMA} we present for reference the performance of a 200-dimensional bi-\texttt{LSTM} classifier trained on full-text. We train the full-text \texttt{LSTM} for 20 epochs across 5 random seeds and select the model with the highest F1-macro performance on the development set. We use the Adam optimizer with a learning rate of $1e-3$ and $1e-5$ weight decay. We report predictive performance and ECE scores on the test-set. 
In Table \ref{tab:F1_ece_hardkuma_STD} we include HardKuma performance with standard deviations, and expected calibration error (ECE), across five runs.

\section{FRESH - Extended
\label{app:extended_fresh}}

Tables \ref{tab:F1_ece_select-then-predict-FRESH_TOPK} and \ref{tab:F1_ece_select-then-predict-FRESH_Contiguous} presents FRESH F1 macro performance and Expected Calibration Error (ECE) for classifiers trained on TopK and Contiguous rationales respectively, with standard deviation in brackets. We include the a priori defined rationale length in the brackets (.\%) and for reference, the ID performance of the Full-Text model (as also seen in Table \ref{tab:F1_ece}). 

Comparing with FRESH performance with Contiguous rationales rather than TopK (see Table \ref{tab:F1_ece_select-then-predict-FRESH_TOPK}), we first observe that performance degrades for most feature attribution methods. These findings are largely in agreement with those of \citet{jain-etal-2020-learning}. However, $x\nabla x$ and IG, which perform poorly with TopK, record surprisingly better scores with Contiguous type rationales. For example, in-domain performance with IG becomes comparable with $\alpha\nabla\alpha$ in in-domain IMDB (83.2 with $\alpha\nabla\alpha$ and 82.5 with IG). This is in sharp contrast with TopK, where IG recorded an F1 score of only 59.7, compared to 87.9 of $\alpha\nabla\alpha$. 

These findings also hold in out-of-domain settings, where $\alpha\nabla\alpha$, $\alpha$ and DeepLift result in poorer FRESH performance with Contiguous type rationales, compared to TopK. However, IG and in many cases $x\nabla x$ improves. For example with TopK rationales, evaluating on Yelp using IG from a model trained on IMDB, results on an F1-score of 69.1. On the contrary, with Contiguous rationales and the same set-up, IG results in FRESH performance of 87.0.

Our findings lead us to assume that, \emph{the rationale type has a large impact on FRESH performance, both in-domain and on out-of-domain settings. Certain feature attribution methods benefit from one type of rationales (e.g. DeepLift with TopK), whilst others from another (e.g. IG with Contiguous).}

\begin{figure*}[!t]
    \centering
    \includegraphics[width=\textwidth]{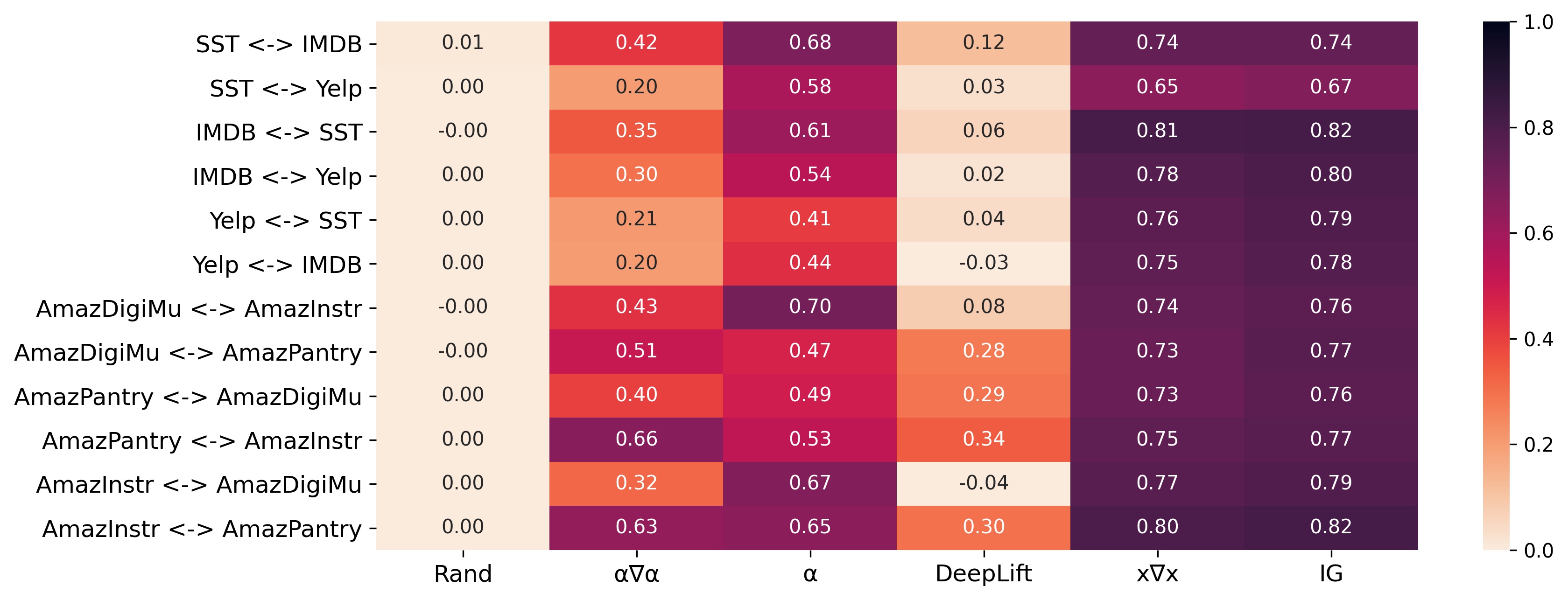}
    \caption{Average Spearman's ranking correlation coefficient, between feature attribution rankings from: (1) a model trained on the same distribution as the evaluation data (ID) and (2) from a model trained in another domain (OOD), such that ID <-> OOD.}
    \label{fig:Spearman_feature_scoring}
\end{figure*}

\section{Extended Analysis
\label{app:extended_correlation_analysis}}

\subsection{Correlation of Rankings}

We examine why $x \nabla x$ and IG, do not perform as well as DeepLift and $\alpha\nabla\alpha$ when using FRESH. We therefore conduct a study to gain better understand this. We first fix the domain of the data we evaluate on and then compute the correlation between importance rankings using any single feature attribution from: (1) a model trained on the same domain with the evaluation data and (2) a model from trained on a different distribution (out-of-domain trained model). High correlations suggest that a feature attribution from an out-of-domain trained model, produce similar importance distributions with that of an in-domain model (i.e. both attend to similar tokens to make a prediction). Therefore, we assume that this will lead to high predictive performance out-of-domain. In Figure \ref{fig:Spearman_feature_scoring} we show Spearman's ranking correlation across dataset pairs, between a model trained on the same distribution as the evaluation data (ID) and an out-of-domain trained model (OOD), such that (ID <-> OOD).
% We hypothesize, that if a model latches on to spurious features from the training distribution, this would not allow it to generalize well. As such, we expect that feature attribution rankings from a model evaluated on in-domain test data to not be correlated with those from a model trained on a different distribution. 
% We consider a strong correlation anything above 0.6, a moderate correlation between 0.40 and 0.60 (including 0.40 and 0.6), and anything below 0.40 as weak to no correlation \citep{akoglu2018user} . 

As expected, the random baseline produced almost no correlation between models. An interesting observation is that two of the gradient-based methods ($x \nabla x$ and IG) produce strongly correlated rankings. This suggests that these two metrics produce generalizable rankings irrespective of the domain shift, when comparing to the remainder of the feature attribution approaches. Surprisingly, Deeplift importance rankings exhibit almost low to no correlation betweenen them, despite being also gradient-based. We hypothesize that this happens because DeepLift considers a baseline input to compute its importance distribution, which highly depends on the model and as such is de-facto normalized and perhaps generalizes better. 
\setlength\tabcolsep{3pt}
\begin{table*}[!t]
    \centering
    \small
    \begin{tabular}{ll||cccccc}
         ID &         OOD & Rand & $\alpha \nabla \alpha$ & $\alpha$ & DeepLift & $x \nabla x$ &    IG \\ \hline
        SST &        IMDB &   0.06 &             0.26 &      0.39 &     0.37 &      0.54 &  0.55 \\
        SST &        Yelp &   0.07 &             0.11 &      0.27 &     0.29 &      0.46 &  0.49 \\
       IMDB &         SST &   0.02 &             0.13 &      0.25 &     0.15 &      0.43 &  0.43 \\
       IMDB &        Yelp &   0.02 &             0.08 &      0.16 &     0.09 &      0.43 &  0.43 \\
       Yelp &         SST &   0.02 &             0.08 &      0.12 &     0.18 &      0.37 &  0.39 \\
       Yelp &        IMDB &   0.02 &             0.05 &      0.12 &      0.10 &       0.40 &  0.41 \\
     AmazDigiMu &   AmazInstr &   0.13 &             0.22 &      0.38 &     0.16 &       0.60 &  0.61 \\
     AmazDigiMu &  AmazPantry &   0.13 &              0.30 &      0.36 &     0.27 &       0.60 &  0.62 \\
     AmazPantry &  AmazDigiMu &   0.14 &             0.28 &      0.35 &     0.27 &       0.60 &  0.63 \\
     AmazPantry &   AmazInstr &   0.14 &             0.39 &      0.42 &     0.21 &      0.62 &  0.64 \\
      AmazInstr &  AmazDigiMu &   0.08 &             0.16 &      0.29 &     0.12 &      0.54 &  0.57 \\
      AmazInstr &  AmazPantry &   0.08 &             0.29 &      0.36 &     0.14 &      0.57 &  0.59 \\
    \end{tabular}
    \caption{Agreement in tokens at 2\% rationale length between a feature attribution from an ID model tested on ID and the same feature attribution trained on an OOD dataset and tested on ID.}
    \label{tab:agreement_on_tokens_@2perc}
\end{table*}

\setlength\tabcolsep{3pt}
\begin{table*}[!t]
    \centering
    \small
    \begin{tabular}{ll||cccccc}
         ID &         OOD & Rand & $\alpha \nabla \alpha$ & $\alpha$ & DeepLift & $x \nabla x$ &    IG \\ \hline
        SST &        IMDB &    0.10 &             0.32 &      0.47 &     0.33 &       0.60 &  0.61 \\
        SST &        Yelp &   0.11 &             0.19 &      0.35 &     0.25 &      0.54 &  0.56 \\
       IMDB &         SST &    0.10 &             0.29 &      0.41 &     0.17 &       0.60 &  0.61 \\
       IMDB &        Yelp &    0.10 &             0.21 &      0.34 &     0.14 &      0.59 &  0.61 \\
       Yelp &         SST &    0.10 &             0.18 &      0.28 &     0.16 &      0.55 &  0.57 \\
       Yelp &        IMDB &    0.10 &             0.16 &      0.29 &     0.12 &      0.56 &  0.58 \\
 AmazDigiMu &   AmazInstr &   0.17 &             0.29 &      0.47 &     0.16 &      0.66 &  0.68 \\
 AmazDigiMu &  AmazPantry &   0.17 &             0.36 &      0.44 &     0.26 &      0.66 &  0.69 \\
 AmazPantry &  AmazDigiMu &   0.17 &             0.33 &      0.42 &     0.27 &      0.66 &  0.68 \\
 AmazPantry &   AmazInstr &   0.17 &             0.46 &      0.49 &     0.24 &      0.67 &  0.69 \\
  AmazInstr &  AmazDigiMu &   0.13 &             0.24 &      0.43 &     0.11 &      0.64 &  0.66 \\
  AmazInstr &  AmazPantry &   0.13 &              0.40 &       0.50 &      0.20 &      0.67 &  0.68 \\
    \end{tabular}
    \caption{Agreement in tokens at 10\% rationale length between a feature attribution from an ID model tested on ID and the same feature attribution trained on an OOD dataset and tested on ID.}
    \label{tab:agreement_on_tokens_@10perc}
\end{table*}
\paragraph{$\boldsymbol\alpha$ for out-of-domain detection?:} An interesting case is that of $\alpha$, where we observe moderate to strong correlations across all test-cases. What is more evident, is that in the OOD tuples we considered, it appears that stronger correlations appear where the OOD task and the ID task are closer together. For example in the case of SST and IMDB (both sentiment analysis tasks for movie reviews), $\alpha$ produces a strong correlation (0.68). This contrasts the moderate correlation of 0.58 between SST and Yelp, which is for restaurant reviews. This is also evident in the case of AmazDigiMu and AmazInstr, where both tasks are for review classification, but for musical related purchases. They both score strong correlations between them and moderate correlations with reviews for pantry purchases (AmazPantry). This observation might suggest, that \emph{using these correlation metrics with $\alpha$ might be an indicator of the degree of task-domain-shift}. Our observation is also supported by the findings of \citet{adebayo-nips}, who show that feature attributions are good indicators of detecting spurious correlation signals in computer vision tasks.Considering $\alpha \nabla \alpha$ we observe a wide range of correlations, ranging from low in the AmazInstr-AmazDigiMu pair to strong in the AmazPantry-AmazInstr pair, which we cannot interpret as something meaningful.

\setlength\tabcolsep{3pt}
\begin{table*}[!t]
    \centering
    \small
    \begin{tabular}{ll||cccccc}
         ID &         OOD & Rand & $\alpha \nabla \alpha$ & $\alpha$ & DeepLift & $x \nabla x$ &    IG \\ \hline
         SST &        IMDB &    0.20 &             0.42 &      0.57 &     0.34 &      0.68 &  0.67 \\
        SST &        Yelp &   0.21 &             0.31 &      0.46 &     0.27 &      0.61 &  0.62 \\
       IMDB &         SST &    0.20 &             0.39 &      0.52 &     0.26 &      0.69 &  0.69 \\
       IMDB &        Yelp &    0.20 &             0.32 &      0.46 &     0.22 &      0.67 &  0.68 \\
       Yelp &         SST &    0.20 &             0.29 &      0.41 &     0.24 &      0.64 &  0.66 \\
       Yelp &        IMDB &    0.20 &             0.27 &      0.42 &      0.20 &      0.65 &  0.66 \\
 AmazDigiMu &   AmazInstr &   0.23 &             0.37 &      0.55 &     0.21 &      0.71 &  0.73 \\
 AmazDigiMu &  AmazPantry &   0.24 &             0.44 &      0.51 &     0.32 &      0.71 &  0.74 \\
 AmazPantry &  AmazDigiMu &   0.24 &              0.40 &       0.50 &     0.33 &      0.71 &  0.73 \\
 AmazPantry &   AmazInstr &   0.24 &             0.54 &      0.57 &     0.32 &      0.72 &  0.73 \\
  AmazInstr &  AmazDigiMu &   0.21 &             0.33 &      0.54 &     0.16 &       0.70 &  0.72 \\
  AmazInstr &  AmazPantry &   0.21 &             0.51 &       0.60 &      0.30 &      0.72 &  0.74 \\
    \end{tabular}
    \caption{Agreement in tokens at 20\% rationale length between a feature attribution from an ID model tested on ID and the same feature attribution trained on an OOD dataset and tested on ID.}
    \label{tab:agreement_on_tokens_@20perc}
\end{table*}

\paragraph{Correlation values and FRESH:}

We first observe that the lowest correlated feature attributions $\alpha \nabla \alpha$ and DeepLift perform the better on FRESH, followed by $\alpha$ which displays moderate correlations and at the end of the spectrum the two gradient-based methods which display high correlations. Contrary to our initial assumption, this suggests that the attributions which generalize better (i.e. return rationales that result in higher FRESH performance) are those which exhibit low to no correlations. 

\paragraph{Agreement at different rationale lengths:} 

As the correlation analysis considers the entire length of the sequence, we now examine a scenario where we have a priori defined rationale lengths. Similarly to the correlation analysis, we now compute the agreement in tokens between ID feature attribution rankings to those of an OOD trained model.  In Tables \ref{tab:agreement_on_tokens_@2perc}, \ref{tab:agreement_on_tokens_@10perc} and \ref{tab:agreement_on_tokens_@20perc} we therefore show the token agreement between in-domain and out-of-domain post-hoc explanations (on the same data) for 2\%, 10\% and 20\% rationale lengths.

Our findings show that across all rationale lengths, results largely agree with the correlation analysis. The two gradient-based methods exhibit higher agreement than the remainder, with $\alpha\nabla\alpha$ and DeepLift recording the lowest agreements. Surprisingly, the poorest performers on out-of-domain FRESH record the highest agreement in tokens with in-domain models. Whilst this suggests that they generalize better, we believe that the inhibiting factor to their performance is their limited in-domain capabilities (i.e. they record the lowest in-domain FRESH performance with TopK).

\section{Post-hoc Explanation Faithfulness - Extended
\label{app:post-hoc-extended}}

In Tables \ref{tab:suff_comp_at_2}, \ref{tab:suff_comp_at_10} and \ref{tab:suff_comp_at_20}, we present post-hoc explanation sufficiency and comprehensiveness scores at 2\%, 10\% and 20\% rationale lengths.

\begin{table*}[!t]
    \centering
    \small
    \begin{tabular}{ll||cccccc||cccccc}
    \textbf{Train} & \textbf{Test} &   \multicolumn{6}{c||}{\textbf{Normalized Sufficiency}} & \multicolumn{6}{c}{\textbf{Normalized Comprehensiveness}}\\
          & & Rand & $\alpha\nabla\alpha$ & $\alpha$ &  DeepLift &       $x\nabla x $ &        IG  & Rand & $\alpha\nabla\alpha$ & $\alpha$ &  DeepLift &       $x\nabla x $ &        IG  \\\hline \hline
                 \multirow{3}{*}{SST} &         SST &       0.42 &                 0.46 &           0.40 &         0.42 &          0.43 &   0.43 &       0.11 &                 0.29 &            0.00 &         0.11 &          0.19 &   0.19 \\ 
                     &        IMDB &       0.35 &                  0.40 &          0.33 &         0.35 &          0.34 &   0.35 &       0.11 &                 0.39 &          0.14 &         0.13 &          0.17 &   0.18 \\
                     &        Yelp &       0.36 &                 0.41 &          0.32 &         0.37 &          0.32 &   0.33 &        0.10 &                 0.31 &          0.08 &          0.10 &          0.11 &   0.13 \\ \hline
                \multirow{3}{*}{IMDB} &        IMDB &       0.36 &                 0.42 &          0.39 &         0.37 &          0.37 &   0.37 &       0.05 &                 0.27 &          0.14 &         0.06 &          0.11 &   0.12 \\
                     &         SST &       0.29 &                  0.30 &          0.29 &          0.30 &           0.30 &    0.30 &       0.16 &                 0.33 &          0.16 &         0.16 &          0.21 &   0.19 \\
                     &        Yelp &        0.40 &                 0.45 &          0.43 &         0.41 &           0.40 &    0.40 &        0.10 &                 0.35 &          0.21 &          0.10 &          0.13 &   0.13 \\ \hline
                \multirow{3}{*}{Yelp} &        Yelp &       0.12 &                 0.13 &          0.13 &         0.13 &          0.13 &   0.13 &       0.02 &                 0.06 &          0.01 &         0.02 &          0.04 &   0.05 \\
                     &         SST &       0.47 &                 0.46 &          0.46 &         0.48 &          0.47 &   0.47 &       0.08 &                 0.09 &            0.00 &         0.09 &          0.12 &   0.12 \\
                     &        IMDB &       0.11 &                 0.11 &          0.11 &         0.12 &          0.11 &   0.11 &       0.07 &                 0.19 &           0.10 &         0.08 &           0.10 &    0.10 \\ \hline
          \multirow{3}{*}{AmazDigiMu} &  AmazDigiMu &       0.24 &                 0.42 &          0.16 &         0.17 &           0.30 &   0.29 &       0.09 &                 0.25 &          0.04 &         0.02 &          0.12 &   0.13 \\
                     &   AmazInstr &       0.17 &                 0.33 &          0.13 &         0.13 &          0.21 &   0.21 &       0.14 &                 0.41 &           0.10 &         0.06 &          0.17 &   0.18 \\
                     &  AmazPantry &       0.27 &                 0.45 &           0.20 &         0.21 &           0.30 &   0.29 &       0.18 &                 0.43 &           0.10 &         0.05 &           0.20 &   0.22 \\ \hline
          \multirow{3}{*}{AmazPantry} &  AmazPantry &       0.23 &                 0.34 &          0.27 &         0.16 &          0.23 &   0.22 &       0.11 &                 0.32 &          0.19 &         0.03 &          0.15 &   0.15 \\
                     &  AmazDigiMu &       0.22 &                 0.35 &          0.29 &         0.16 &          0.22 &   0.22 &        0.10 &                 0.29 &          0.19 &         0.03 &          0.12 &   0.12 \\
                     &   AmazInstr &       0.14 &                 0.23 &          0.18 &         0.11 &          0.15 &   0.14 &       0.12 &                 0.39 &          0.23 &         0.07 &          0.16 &   0.17 \\ \hline
           \multirow{3}{*}{AmazInstr} &   AmazInstr &       0.13 &                 0.18 &          0.09 &         0.11 &          0.13 &   0.13 &       0.16 &                  0.40 &          0.05 &         0.08 &          0.17 &   0.18 \\
                     &  AmazDigiMu &       0.19 &                 0.29 &          0.12 &         0.13 &          0.19 &   0.18 &       0.14 &                 0.35 &          0.04 &         0.05 &          0.14 &   0.15 \\
                     &  AmazPantry &        0.20 &                  0.30 &          0.14 &         0.15 &           0.20 &    0.20 &       0.19 &                 0.45 &          0.04 &         0.08 &          0.18 &   0.21 \\
    \end{tabular}
    \caption{Normalized Sufficiency and Comprehensiveness (higher is better) in-domain and out-of-domain at 2\% rationale length, for five feature attribution approaches and a random attribution baseline.}
    \label{tab:suff_comp_at_2}
\end{table*}

\begin{table*}[!t]
    \centering
    \small
    \begin{tabular}{ll||cccccc||cccccc}
    \textbf{Train} & \textbf{Test} &   \multicolumn{6}{c||}{\textbf{Normalized Sufficiency}} & \multicolumn{6}{c}{\textbf{Normalized Comprehensiveness}}\\
          & & Rand & $\alpha\nabla\alpha$ & $\alpha$ &  DeepLift &       $x\nabla x $ &        IG  & Rand & $\alpha\nabla\alpha$ & $\alpha$ &  DeepLift &       $x\nabla x $ &        IG  \\\hline \hline
               \multirow{3}{*}{SST} &         SST &       0.43 &                 0.55 &          0.43 &         0.46 &          0.44 &   0.45 &       0.16 &                 0.42 &           0.20 &         0.22 &          0.25 &   0.25 \\
                 &        IMDB &       0.36 &                 0.65 &          0.44 &         0.37 &          0.36 &   0.36 &       0.19 &                 0.69 &          0.39 &         0.24 &          0.25 &   0.26 \\
                 &        Yelp &       0.37 &                 0.67 &          0.37 &         0.39 &          0.33 &   0.34 &       0.17 &                 0.58 &          0.25 &          0.20 &          0.22 &   0.24 \\ \hline
            \multirow{3}{*}{IMDB} &        IMDB &       0.37 &                 0.64 &          0.54 &          0.40 &          0.39 &   0.39 &        0.10 &                 0.55 &           0.30 &         0.17 &          0.18 &   0.18 \\
                 &         SST &       0.28 &                 0.32 &          0.29 &          0.30 &           0.30 &    0.30 &       0.23 &                 0.48 &          0.29 &         0.29 &           0.30 &   0.29 \\
                 &        Yelp &       0.41 &                 0.54 &          0.46 &         0.43 &          0.41 &   0.41 &       0.18 &                 0.58 &          0.36 &         0.22 &          0.24 &   0.24 \\ \hline
            \multirow{3}{*}{Yelp} &        Yelp &       0.17 &                 0.22 &          0.23 &         0.26 &          0.19 &    0.20 &       0.05 &                 0.15 &          0.05 &         0.06 &          0.08 &   0.08 \\
                 &         SST &       0.48 &                 0.49 &          0.47 &          0.50 &          0.46 &   0.46 &       0.13 &                 0.23 &          0.15 &         0.16 &           0.20 &    0.20 \\
                 &        IMDB &       0.13 &                 0.29 &          0.29 &         0.22 &          0.14 &   0.15 &       0.13 &                 0.35 &          0.28 &         0.16 &          0.18 &   0.19 \\ \hline
      \multirow{3}{*}{AmazDigiMu} &  AmazDigiMu &       0.33 &                 0.67 &          0.24 &         0.25 &          0.39 &   0.36 &       0.11 &                 0.34 &          0.08 &         0.06 &          0.15 &   0.16 \\
                 &   AmazInstr &       0.28 &                 0.67 &          0.22 &         0.26 &          0.29 &   0.28 &       0.19 &                 0.57 &          0.19 &         0.15 &          0.22 &   0.24 \\
                 &  AmazPantry &       0.33 &                 0.64 &          0.25 &         0.28 &          0.36 &   0.34 &       0.22 &                 0.55 &          0.17 &         0.12 &          0.25 &   0.26 \\ \hline
      \multirow{3}{*}{AmazPantry} &  AmazPantry &       0.23 &                 0.46 &          0.34 &         0.17 &          0.24 &   0.23 &       0.15 &                 0.45 &          0.29 &          0.10 &           0.20 &   0.21 \\
                 &  AmazDigiMu &       0.22 &                 0.46 &          0.35 &         0.16 &          0.23 &   0.22 &       0.13 &                 0.42 &          0.29 &          0.10 &          0.17 &   0.17 \\
                 &   AmazInstr &       0.14 &                 0.42 &          0.27 &         0.12 &          0.16 &   0.15 &       0.18 &                 0.59 &           0.40 &         0.17 &          0.24 &   0.25 \\ \hline
       \multirow{3}{*}{AmazInstr} &   AmazInstr &       0.13 &                 0.28 &          0.09 &         0.12 &          0.13 &   0.13 &       0.23 &                 0.58 &          0.16 &         0.22 &          0.24 &   0.25 \\
                 &  AmazDigiMu &       0.19 &                 0.32 &          0.12 &         0.14 &           0.20 &   0.18 &       0.18 &                 0.47 &           0.10 &         0.14 &           0.20 &    0.20 \\
                 &  AmazPantry &       0.21 &                 0.35 &          0.15 &         0.17 &          0.21 &   0.21 &       0.24 &                 0.57 &          0.12 &         0.18 &          0.24 &   0.27 \\
    \end{tabular}
    \caption{Normalized Sufficiency and Comprehensiveness (higher is better) in-domain and out-of-domain at 10\% rationale length, for five feature attribution approaches and a random attribution baseline.}
    \label{tab:suff_comp_at_10}
\end{table*}

\begin{table*}[!t]
    \centering
    \small
    \begin{tabular}{ll||cccccc||cccccc}
    \textbf{Train} & \textbf{Test} &   \multicolumn{6}{c||}{\textbf{Normalized Sufficiency}} & \multicolumn{6}{c}{\textbf{Normalized Comprehensiveness}}\\
          & & Rand & $\alpha\nabla\alpha$ & $\alpha$ &  DeepLift &       $x\nabla x $ &        IG  & Rand & $\alpha\nabla\alpha$ & $\alpha$ &  DeepLift &       $x\nabla x $ &        IG  \\\hline \hline
                 \multirow{3}{*}{SST} &         SST &       0.45 &                 0.68 &          0.51 &         0.51 &          0.48 &   0.49 &       0.22 &                 0.54 &          0.34 &         0.33 &          0.32 &   0.34 \\
                 &        IMDB &       0.38 &                 0.77 &          0.55 &         0.39 &          0.37 &   0.38 &       0.29 &                  0.80 &          0.54 &         0.36 &          0.34 &   0.36 \\
                 &        Yelp &       0.39 &                 0.83 &          0.57 &         0.41 &          0.37 &   0.38 &       0.25 &                 0.71 &          0.44 &          0.30 &          0.32 &   0.34 \\ \hline
            \multirow{3}{*}{IMDB} &        IMDB &       0.37 &                 0.75 &          0.62 &         0.42 &          0.41 &   0.42 &       0.16 &                 0.73 &          0.47 &          0.30 &          0.27 &   0.27 \\
                 &         SST &       0.26 &                  0.40 &          0.31 &         0.31 &          0.31 &    0.30 &       0.32 &                 0.65 &          0.42 &         0.41 &          0.42 &   0.42 \\
                 &        Yelp &       0.42 &                 0.62 &           0.50 &         0.43 &          0.44 &   0.44 &       0.28 &                 0.67 &          0.47 &         0.35 &          0.36 &   0.37 \\ \hline
            \multirow{3}{*}{Yelp} &        Yelp &       0.25 &                 0.43 &          0.41 &          0.40 &          0.28 &    0.30 &       0.09 &                 0.25 &          0.12 &         0.13 &          0.14 &   0.15 \\
                 &         SST &       0.49 &                 0.55 &          0.51 &         0.53 &          0.48 &   0.48 &        0.20 &                 0.35 &          0.27 &         0.26 &          0.28 &   0.29 \\
                 &        IMDB &       0.19 &                 0.53 &           0.50 &         0.34 &          0.24 &   0.25 &        0.20 &                 0.46 &           0.40 &         0.27 &          0.28 &   0.28 \\ \hline
      \multirow{3}{*}{AmazDigiMu} &  AmazDigiMu &       0.43 &                 0.81 &          0.47 &         0.35 &          0.52 &    0.50 &       0.14 &                 0.41 &          0.17 &          0.10 &          0.19 &    0.20 \\
                 &   AmazInstr &       0.37 &                 0.79 &          0.49 &         0.42 &          0.43 &   0.42 &       0.24 &                 0.63 &          0.33 &         0.23 &          0.28 &    0.30 \\
                 &  AmazPantry &       0.42 &                 0.76 &          0.45 &         0.37 &          0.47 &   0.45 &       0.26 &                 0.61 &          0.31 &          0.20 &           0.30 &   0.32 \\ \hline
      \multirow{3}{*}{AmazPantry} &  AmazPantry &       0.27 &                 0.63 &          0.46 &         0.19 &           0.30 &   0.29 &       0.21 &                 0.57 &           0.40 &         0.17 &          0.28 &   0.29 \\
                 &  AmazDigiMu &       0.25 &                 0.63 &          0.46 &         0.18 &          0.28 &   0.27 &       0.19 &                 0.55 &          0.39 &         0.16 &          0.25 &   0.25 \\
                 &   AmazInstr &       0.16 &                 0.61 &          0.42 &         0.14 &          0.21 &    0.20 &       0.27 &                 0.72 &          0.54 &         0.26 &          0.35 &   0.36 \\ \hline
       \multirow{3}{*}{AmazInstr} &   AmazInstr &       0.15 &                 0.46 &          0.15 &         0.18 &          0.17 &   0.16 &       0.31 &                 0.72 &          0.33 &         0.34 &          0.32 &   0.34 \\
                 &  AmazDigiMu &       0.21 &                 0.46 &          0.16 &         0.17 &          0.23 &    0.20 &       0.24 &                  0.60 &          0.22 &         0.22 &          0.26 &   0.27 \\
                 &  AmazPantry &       0.23 &                 0.49 &          0.18 &         0.21 &          0.24 &   0.23 &       0.31 &                 0.68 &          0.28 &         0.28 &          0.32 &   0.35 \\
    \end{tabular}
    \caption{Normalized Sufficiency and Comprehensiveness (higher is better) in-domain and out-of-domain at 20\% rationale length, for five feature attribution approaches and a random attribution baseline.}
    \label{tab:suff_comp_at_20}
\end{table*}

\end{document}